%% file: automatikz.tex
\pdfoutput=1
\PassOptionsToPackage{dvipsnames,table}{xcolor}
\PassOptionsToPackage{outputdir=build,cachedir=minted,frozencache}{minted}
\documentclass{article} 
\usepackage{styles/iclr2024_conference}
\usepackage{styles/automatikz} 
\usepackage{hyperref}
\usepackage{graphicx}
\usepackage{enumitem}
\usepackage{csquotes}
\usepackage{tikz,pgfplots,calc}
\usepackage{booktabs,wrapfig,subcaption}
\usepackage[most,minted]{tcolorbox}
\usetikzlibrary{calc}
\pgfplotsset{compat=1.18}

\graphicspath{{figures}{figures/clima}{figures/examples}{figures/python}{build/figures}}
\makeatletter\let\input@path\Ginput@path\makeatother

\iclrfinalcopy%

\hypersetup{%
  colorlinks=true,
  citecolor=darkblue,
  linkcolor=darkblue,
  urlcolor=darkblue
}

\setlist[enumerate]{label= (\roman*)}

\title{\projectname: Text-Guided Synthesis of Scientific Vector Graphics with
\tikzname}

\author{%
  Jonas Belouadi\\
  \href{https://nl2g.github.io}{\normalcolor Natural Language Learning Group}\\
  Bielefeld University, Germany\\
  \mail{jonas.belouadi@uni-bielefeld.de}
  \And%
  Anne Lauscher\\
  \href{https://www.bwl.uni-hamburg.de/en/ds}{\normalcolor Data Science Group}\\
  University of Hamburg, Germany\\
  \mail{anne.lauscher@uni-hamburg.de}
  \AND%
  Steffen Eger\\
  \href{https://nl2g.github.io}{\normalcolor Natural Language Learning Group}\\
  University of Mannheim, Germany\\
  \mail{steffen.eger@uni-mannheim.de}
}

\begin{document}

\maketitle\ificlrfinal\else\vspace{83pt}\fi

\begin{abstract}
  Generating bitmap graphics from text has gained considerable attention, yet
  for scientific figures, vector graphics are often preferred. Given that
  vector graphics are typically encoded using low-level graphics primitives,
  generating them directly is difficult.
  To address this, we propose the use of \tikzname, a well-known abstract
  graphics language that can be compiled to vector graphics, as an intermediate
  representation of scientific figures.\ \tikzname offers human-oriented,
  high-level commands, thereby facilitating conditional language modeling with
  any large language model.
  To this end, we introduce \dataset, the first large-scale \tikzname dataset
  consisting of 120k \tikzname drawings aligned with captions. We fine-tune
  \llama on \dataset, as well as our new model \clima, which augments \llama
  with multimodal \clip embeddings. In both human and automatic evaluation,
  \clima and \llama outperform commercial \gpt and \claude in terms of
  similarity to human-created figures, with \clima additionally improving
  text-image alignment.
  Our detailed analysis shows that all models generalize well and are not
  susceptible to memorization.\ \gpt and \claude, however, tend to generate
  more simplistic figures compared to both humans and our models.
  We make our framework, \projectname, along with model weights and datasets,
  publicly
  available.\footnote{\anonymize{\url{https://github.com/potamides/AutomaTikZ}}}
\end{abstract}

\input{sections/introduction}
\input{sections/related}
\input{sections/data}
\input{sections/methods}
\input{sections/experiments}
\input{sections/analysis}
\input{sections/conclusion}
\input{sections/ethics}
\anonymize[]{\input{sections/acknowledgements}}

\bibliography{automatikz}
\bibliographystyle{iclr2024_conference}

\clearpage
\appendix
\input{sections/appendix}

\end{document}

%% file: sections/introduction.tex
\section{Introduction}\label{sec:introduction} 
Recent advancements in text-to-image generation have facilitated the generation
of detailed images from simple natural language
descriptions~\citep{esser2021vqgan,ramesh2021dalle,ramesh2022dalle2,saharia2022photorealistic,rombach2022highresolution,zhang2023adding}.
Models like \stablediffusion~\citep{rombach2022highresolution} and
\dalle~\citep{ramesh2021dalle,ramesh2022dalle2} often yield results comparable
to real photographs or human-created artworks.
However, these models primarily generate \emph{raster graphics}, typically at
low resolutions, which are not ideal for \emph{scientific figures}. Researchers
use scientific figures to convey complex ideas or present critical findings,
making them central to scientific
research~\citep{tufte1992visual,hsu2021scicap}. Consequently, they demand a
high degree of geometric precision and legible text, even at small font sizes,
areas where raster graphics fall short. As a result, many research conferences
advocate the use of \emph{vector
graphics},\footnote{\url{https://acl-org.github.io/ACLPUB/formatting.html}}
which decompose information into geometric shapes, allow searchable text, and
usually have smaller file sizes.

Automated vector graphics generation is a growing research area as
well~\citep{lopes2019svgvae,carlier2020deepsvg,aoki2022chinese,ma2022live,frans2022clipdraw,jain2023vectorfusion,wu2023iconshop},
but current methods have their own share of limitations. Specifically, they
mainly generate low-level path elements of the Scalable Vector Graphics~(SVG)
format, either failing to maintain accurate geometric
relations~\citep{ma2022live,frans2022clipdraw,jain2023vectorfusion} or only
generating outputs of limited complexity such as single icons or font
characters~\citep{lopes2019svgvae,carlier2020deepsvg,aoki2022chinese,wu2023iconshop}.

\begin{figure}
  \centering
  \input{clima.tex}
  \caption{Exemplary scientific figures generated with \clima. \clima takes the
  captions as input, processes them with \clip and \llama, and
  generates \tikzname drawings that compile to vector graphics.}%
  \label{fig:clima}
\end{figure}
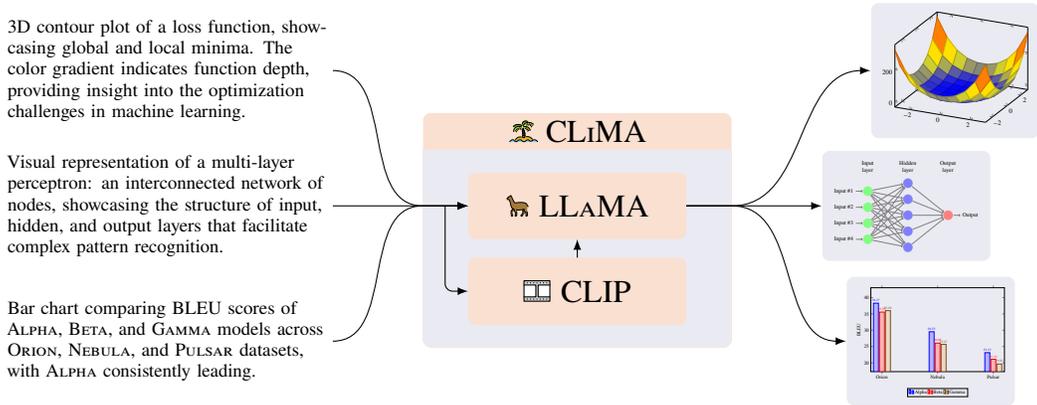

To address these limitations, we explore the use of \emph{graphic languages},
which abstract from lower-level vector graphics formats by providing high-level
constructs that can be compiled to such
formats~\Citep{zandt2007pstricks,hobby2014metapost,tantau2023tikz}.
Language models show potential in learning these languages to solve simple
tasks~\citep{bubeck2023sparks,zhang2023controllable}, but the depth of this
capability, i.e., whether it can produce scientific figures, remains
unexplored. Due to its expressiveness and emphasis on science, which enables
the creation of complex figures with only a few commands, we focus on the
graphics language \emph{\tikzname} in this work~\citep{tantau2023tikz}. We aim
to understand whether language models can capture the nuances of \tikzname and
automatically generate scientific figures based on image captions, analogous to
text-to-image generation. This could not only enhance productivity and foster
inclusiveness~(aiding researchers less versed in programming-like languages,
such as social scientists), but also aid education by creating tailored
\tikzname examples. The use case for this is demonstrated by the \tex Stack
Exchange\footnote{\url{https://tex.stackexchange.com}}, where nearly 10\% of
the asked questions pertain to \tikzname, making it the most frequently
discussed topic on the platform. Our key contributions are as follows:
\begin{enumerate}
  \item As part of our \projectname project, we create \dataset, the first
    large-scale \tikzname dataset to our knowledge, featuring approximately
    120k paired \tikzname drawings and captions.
  \item We fine-tune the large language model~(LLM)
    \llama~\citep{touvron2023llama} on \dataset and compare its performance to
    general-purpose LLMs, specifically \gpt~\citep{openai2023gpt4} and
    \claude~\citep{anthropic2023claude}. Both automatic and human evaluation
    agree that scientific figures generated by fine-tuned \llama resemble
    human-created figures more closely.
  \item We further develop \clima, a variant of \llama augmented with
    multimodal \clip embeddings~(cf.\ \figref{fig:clima};
    \citealp{radford2021learning}). This enhancement allows \clima to visually
    interpret input captions, thereby improving text-image alignment. It also
    enables the use of images as supplementary inputs, leading to a further
    boost in performance.
  \item In addition, we demonstrate that all models exhibit few memorization
    problems and generate novel outputs. However, \gpt and \claude tend to
    generate simpler outputs than \llama and \clima, sometimes producing
    degenerate solutions that maximize text-image similarity by visibly copying
    the input caption into the output image.
\end{enumerate}

%% file: figures/clima.tex
\begingroup

\usetikzlibrary{calc}
\newsavebox{\contour}
\newsavebox{\barchart}
\newsavebox{\mlp}
\sbox{\contour}{\input{contour.tex}}
\sbox{\barchart}{\input{barchart.tex}}
\sbox{\mlp}{\input{mlp.tex}}

\newtcolorbox{model}[2][]{
  enhanced jigsaw,
  fonttitle=\large,
  fontupper=\large,
  colback=examplebg,
  colframe=exampletitle,
  coltitle=black,
  halign=center,
  valign=center,
  center title,
  boxrule=0pt,
  left=5mm,
  right=5mm,
  title={#2},
  #1
}

\newtcolorbox{component}[1][]{
  enhanced,
  colback=exampletitle,
  boxrule=0pt,
  halign=center,
  valign=center,
  frame hidden,
  #1
}

\begin{tikzpicture}[%
  remember picture,
  x=1.3cm,y=1.8cm,
  font=\scriptsize,
  model/.style={inner sep=0pt},
  Caption/.style={anchor=east},
  Figure/.style={fill=examplebg,anchor=west,rounded corners=1mm},
]

  \node[model] (clima) {
    \begin{model}[width=4.1cm]{\includegraphics[height=\heightof{\clima}]{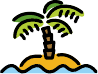} \clima}
      \begin{component}[remember as=llama]
        \includegraphics[height=\heightof{\llama}]{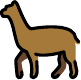}
        \llama
      \end{component}
      \begin{component}[remember as=clip]
        \includegraphics[height=\heightof{\clip}]{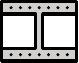}
        \clip
      \end{component}
    \end{model}
  };

  \node[Caption] at ($(llama)+(-2.5,+1)$)  (caption1) [text width=4.2cm] {%
    3D contour plot of a loss function, showcasing global and local minima.
    The color gradient indicates function depth, providing insight into the
    optimization challenges in machine learning.
  };
  \node[Caption] at ($(llama)+(-2.5, 0)$)  (caption2) [text width=4.2cm] {%
    Visual representation of a multi-layer perceptron: an interconnected
    network of nodes, showcasing the structure of input, hidden, and output
    layers that facilitate complex pattern recognition.
  };
  \node[Caption] at ($(llama)+(-2.5,-1)$)  (caption3) [text width=4.2cm] {%
    Bar chart comparing \bleu scores of \textsc{Alpha}, \textsc{Beta}, and
    \textsc{Gamma} models across \textsc{Orion}, \textsc{Nebula}, and
    \textsc{Pulsar} datasets, with \textsc{Alpha} consistently leading.
  };

  \node[Figure] at ($(llama)+(3,    1)$) (figure1) {\resizebox{2cm}{!}{\usebox{\contour}}};
  \node[Figure] at ($(llama)+(2.5,  0)$) (figure2)  {\resizebox{2cm}{!}{\usebox{\mlp}}};
  \node[Figure] at ($(llama)+(2.75,-1)$) (figure3) {\resizebox{2cm}{!}{\usebox{\barchart}}};

  \coordinate (in)   at (clima.west |- llama);
  \coordinate (fork) at ($(clima.west |- llama)+(3mm,0)$);
  \coordinate (out)  at (clima.east |- llama);

  \foreach \index in {1,2,3} \draw[->,>=latex] (caption\index.east) to [out=0,in=180] (in) to (llama.west);
  \foreach \index in {1,2,3} \draw[->,>=latex] (llama.east) to (out) to [out=0,in=180] (figure\index.west);

  \draw[->,>=latex] (clip.north) to (llama.south);
  \draw[->,>=latex,rounded corners=1mm] (fork) |- (clip.west);
\end{tikzpicture}

\endgroup

%% file: sections/related.tex
\section{Related Work}\label{sec:related-work} 
Our work connects to several distinct but interrelated fields, namely
text-to-image generation, vector graphics generation, scientific figure
understanding, and code generation. For each field, we provide a comprehensive
review of the most relevant prior work.

\paragraph{Text-to-Image \& Vector Graphics Generation}
The evolution of text-to-image generation can be characterized by three
development stages: Generative Adversarial
Networks~\citep{reed2016generative,zhang2017stackgan,brock2019large,kang2023scaling},
auto-regressive
models~\citep{ramesh2021dalle,ding2021cogview,ding2022cogview2,chang2023muse},
and diffusion
models~\citep{rombach2022highresolution,ramesh2022dalle2,saharia2022photorealistic,zhang2023adding}.
Although \citet{rodriguez2023figgen,rodriguez2023ocrvqgan} explore their use
for scientific figures, these approaches are inherently limited to generating
raster graphics.

Vector graphics generation has evolved as a parallel field. Building upon
innovative work in sketch generation~\citep{ha2018neural},
\citet{lopes2019svgvae} generate single SVG font characters made up of straight
lines and Bézier curves.\ \citet{carlier2020deepsvg} extend this approach to
include SVG \emph{icons}. However, none of these models support text
conditioning.
Another branch of research focuses on vectorizing text-to-image
models~\citep{ma2022live,frans2022clipdraw,jain2023vectorfusion}. Although
these methods enable text conditioning, text-to-image models typically have
difficulties producing flat-colored SVG-style images, and the vectorized
results tend to have imprecise geometric relations and jagged
paths~\citep{wu2023iconshop}.
Addressing these challenges, \citet{cai2023leveraging} and
\citet{wu2023iconshop} investigate auto-regressive language modeling directly
on SVG representations. Even though these approaches better capture the
aesthetics of vector graphics, they are limited to editing existing graphics or
generating monochrome icons of limited complexity.

\paragraph{Scientific Figure Understanding}
Despite the limited number of approaches dedicated to generating
scientific figures, scientific figure \emph{understanding} is a subject of
extensive research. Arguably, the task that is inverse to ours is the
captioning of scientific figures. Expanding on prior work in scientific visual
question-answering~\citep{siegel2016figureseer,kahou2018figureqa,dvqa2018kafle},
\citet{chen2019caption,chen2019figure,chen2020relmaps} train a captioning model
using a corpus of synthesized figure-caption pairs.\ \citet{hsu2021scicap}
extend this approach to real scientific figures, noting substantial challenges.
To improve performance, \citet{yang2023scicap} reformulate the task, augmenting
captions with OCR tokens and paragraphs referencing the figures.\
\citet{singh2023figcapshf} take a different approach and utilize reinforcement
learning to consider the quality of the captions during training.
In addition to such task-oriented models, recent advancements in multimodal
large language modeling~(MLLM\@;
\citealp{liu2023visual,dai2023instructblip,yin2023survey}) allow for
generalized visual reasoning about scientific
figures~\citep{ye2023mplugdocowl,zhang2023llavar,horawalavithana2023scitune}.

\paragraph{Code Generation}
As graphics languages are a subset of programming languages, our work is
closely related to code generation~\citep{xu2022codeeval}. At its core is the
ongoing research on pre-training or fine-tuning LLMs on
code~\citep{roziere2023code,li2023starcoder,fried2023incoder,li2022alphacode,chen2021evaluating},
commonly with a multitask objective~\citep{fried2023incoder} consisting of
causal language modeling and infilling
prediction~\citep{bavarian2022efficient}. Despite the significant amount of
recent progress, the primary focus of code generation remains on high-resource
programming languages such as Python, Java, or
JavaScript~\citep{zan2023lnlcode}.\ \tikzname commands, in comparison, are
invoked as \tex macros, and the \tex programming language is considered
low-resource and typically overlooked in model evaluations. Yet, \tex may still
exist in training corpora, as evidenced by \gpt's ability to comprehend \tex
and \tikzname~\citep{bubeck2023sparks,zhang2023controllable}. As far as we
know, there has been no comprehensive evaluation of this capability, which we
also address in this work.

%% file: sections/data.tex
\section{The \dataset Dataset} 
\dataset is, to our best knowledge, the first large-scale dataset of \tikzname
drawings with corresponding captions.\ \tikzname is well-known within the \tex
community, but its resources are scattered across the internet, making the
creation of a large dataset a fundamental challenge of our work. Consequently,
we gather \tikzname drawings and their captions from a variety of online
resources, as detailed below.

\subsection{Data Acquisition}\label{sec:data-acquisition}
We collect the data from dedicated websites, online repositories, the \tex
Stack Exchange, \arxiv papers, and also artificial examples. A comprehensive
overview of our data collection is provided in \tabref{tab:dataset}. We gather
\tikzname drawings created between November 2006 and June 2023 that
successfully compile with \texlive.\footnote{\url{https://tug.org/texlive}}
Ablation studies and examples can be found in
\appappref{sec:additional-experiments}{sec:examples}.
\begin{wraptable}[11]{R}{0pt}
  \centering
  \begin{tabular}{lr d{2.2}<{\rlap{\%}}}
    \toprule
    \thead{Source} & \nhead{Size} & \nhead{Augmented}\\
    \midrule
    Curated Examples     &      981 & 63.2\\
    \tex Stack Exchange  &  29\,238 & 51.31\\
    \Arxiv Papers        &  85\,656 & 67.75\\
    Artificial Examples  &   3\,914 & 50\\
    \midrule
    All                  & 119\,789 & 62.71\\
    \bottomrule
  \end{tabular}
  \caption{Detailed breakdown of \dataset showing size and percentage of
  augmented data for the whole dataset and each source individually.}%
  \label{tab:dataset}
\end{wraptable}

\paragraph{Curated Examples}
Several websites and online repositories\footnote{\url{https://texample.net},
\url{https://tikz.net}, \url{https://pgfplots.net} \& \url{https://github.com}
projects} focus on collecting and sharing \tikzname drawings for educational
purposes and to showcase high-quality examples. Through web scraping, we
retrieve any \tikzname drawings from these sites that have associated captions.

\paragraph{\tex Stack Exchange}
We also source \tikzname drawings from \tex Stack Exchange~(cf.\
\secref{sec:introduction}). We examine the quarterly data dump and extract
questions tagged with \tikzname and relevant answers with a minimum score of 1.
To convert textual questions into image captions, we utilize
\wizardlm~\citep{xu2023wizardlm}, an LLM trained to follow arbitrary
instructions. Using the title and body of a question as context, we task
\wizardlm with creating a descriptive caption for the figure provided in the
answer. More details on the caption generation procedure can be found in
\appref{sec:prompting}.

\paragraph{\Arxiv Papers}
\Arxiv\footnote{\url{https://arxiv.org}} is a widely-used open-access archive
for scientific articles. As \arxiv encourages researchers to upload their
papers alongside their source files, it serves as a valuable resource for
obtaining \tikzname drawings. Initially, we retrieve all papers with \tex
source files and retain those that use the \tikzname package. Subsequently, we
expand any include directives and extract all \tikzname environments using
regular expressions. To ensure compilability, we additionally preserve all
required preamble elements. For that, we first establish a set of rules by
analyzing documents obtained from other sources that determine which package
imports and configuration options should be retained. We then parse all macro
definitions and keep for each \tikzname drawing the macros it uses. Finally, we
exclude any \tikzname drawings that fail to compile after this extraction
process~(around 120k).

\paragraph{Artificial Examples}
\gpt has demonstrated the emergent ability to generate simple, tangible
objects~(e.g., unicorns) in \tikzname~\citep{bubeck2023sparks}. While not the
primary focus of this work, we seek to transfer this ability to our models
through knowledge distillation~\citep{bucila2006compression}. To this end, we
compile a diverse set of object descriptions derived from the object categories
in the \mscoco, \lvis, and \visor
datasets~\citep{lin2014microsoft,gupta2019lvis,gokhale2023benchmarking}.
Moreover, we sample emoji descriptions from the \openmoji
database.\footnote{\url{https://openmoji.org}} Following this, we instruct \gpt
to generate a \tikzname drawing for each description, using a chain-of-thought
prompt~\citep{wei2023chainofthought} we adopt from
\citet{zhang2023controllable}, as detailed in \appref{sec:prompting}.

\subsection{Data Augmentation}\label{sec:data-augmentation}
Prior research indicates a correlation between caption length and caption
quality~\citep{gelman2002preach,hartley2003single,huang2023summaries}. Notably,
\citet{huang2023summaries} propose a heuristic to judge scientific captions
with less than 30 tokens as being of poor quality.
Given the recent advancements in MLLM, and notably in MLLM with a focus on
science~(cf.\ \secref{sec:related-work}), we propose the automatic augmentation
of such captions~\citep{belouadi2023bygpt5,belouadi2023uscore}. For an in-depth
analysis of implications, refer to \appref{sec:additional-experiments}.

Specifically, we leverage \llavar~\citep{zhang2023llavar}, instructing it to
generate short descriptions for \tikzname drawings with captions containing
fewer than 30 tokens~(cf.\ \appref{sec:prompting}).\footnote{In this work, we
use the \moses tokenizer~\citep{koehn2007moses} to count tokens.}
Inspired by the \capfilt method~\citep{li2022blip}, we generate five candidate
descriptions and rank them based on their text-image similarity using
\clipscore~\citep{hessel2021clipscore}. The final augmented caption is then
formed by concatenating the original caption with the top-ranked description.
For \gpt, we cannot rely on the heuristic, as the captions used are not
scientific. Instead, we augment all captions to increase diversity in our
dataset while retaining the original captions as well.
\tabref{tab:dataset} displays the percentage of augmented captions in our
dataset. On average, this method increases the \clipscore of captions with
originally fewer than 30 tokens from 24.76 to 29.12, a substantial improvement
in text-image similarity, especially considering that \clipscore typically
ranges between zero and 40~\citep{hessel2021clipscore}. The \clipscore for
original captions exceeding 30 tokens is 27.06.

%% file: sections/methods.tex
\section{Methods}\label{sec:methods} 
We leverage \llama~\citep{touvron2023llama} as the base model in most
experiments, using captions from \dataset as model input and \tikzname code as
ground truths. Since \tex source files from \arxiv were included in
\llama's pre-training data, it may have prior knowledge beneficial for this
task. We choose the original \llama release over its updated revisions,
\llama[2]~\citep{touvron2023llama2} and \codellama~\citep{roziere2023code}, as
their training data is not as clearly specified. This uncertainty and
their more recent release would make it difficult to create a test set without
training-to-test data leakage.
We also experiment with \gpt and \claude, as earlier research hints at inherent
potential for our task~(cf.\ \secref{sec:data-acquisition} and
\secref{sec:related-work}), and employ the same chain-of-thought approach
outlined in \secref{sec:data-acquisition}. However, as they are proprietary, we can
only address data leakage for
\llama~\citep{aiyappa2023trust}.

\subsection{\clima}\label{sec:clima}
A potential drawback of vanilla \llama is that it may not understand visual
concepts, given that it was not designed to process image data. However, this
ability could significantly enhance the creation of scientific figures.
Therefore, we modify \llama by combining it with a \clip \vit
model~\citep{cherti2023clip}.\ \clip is frequently employed to establish a
bridge between vision and natural language, thereby facilitating the creation
of MLLMs~\citep{yin2023survey}.

However, unlike most MLLM methods, to our knowledge we are the first to use
\clip's \emph{multimodal} projection layer, which allows us to extract visual
information from both text and images in a common embedding space~(cf. \
\figref{fig:clima}). This approach is akin to text-to-image models like \dalle
and \clipgen~\citep{wang2022clipgen}, that make use of this duality to generate
raster graphics.
In our case, our primary objective is to provide \llama with a visual
interpretation of the input caption, anticipating that this adjustment will
boost the alignment with generated \tikzname drawings. In addition, it also
enables us to experiment with supplying rasterized scientific figures as an
additional input~(cf.\ \secref{sec:experiments}). As this new model can be
described as using \textsc{\textbf{CL}IP} \textbf{i}nside
\textsc{LLa\textbf{MA}}, we refer to it as \emph{\clima}.

We accomplish this integration by connecting \clip's output with \llama's input
via soft prompting~\citep{lester2021power}; i.e., we prepend \clip's embedding
vector to \llama's input embeddings of the caption. This requires adding a
feed-forward layer with dimensions $\delta_{\vit}\times\delta_{\llama}$ to
connect image features of dimension $\delta_{\vit}$ with \llama's word
embedding space of dimension $\delta_{\llama}$. Following insights from
\citet{liu2023visual}, we pre-train this adaption layer on a dataset of 595K
generic text-image pairs for one epoch while keeping both \clip and \llama
frozen during the process.

\subsection{Error Handling \& Correction}\label{sec:error-handling}
A general issue with our language modeling approach to \tikzname generation is
that outputs may violate the syntactic and semantic rules of \tex, potentially
leading to errors and uncompilable documents.
While there are constrained decoding algorithms that can force models to form
valid programs~\citep{poesia2022synchromesh,scholak2021picard}, they depend on
parse trees and are only useful for languages with a context-free grammar.\
\tex, however, has a flat, unstructured syntax that is generally impossible to
parse~\citep{erdweg2010tex}, rendering these methods unsuitable for our
approach.

As an alternative, we propose an \emph{iterative resampling} method, leveraging
the diagnostic data produced during compilation. If an error arises during
compilation, we analyze the logfile to identify its source. Rather than
resampling from the start, we then reverse the generation process to just
before the error line and continue sampling from there. If the error persists,
we infer that the origin of the problem lies earlier in the code and reverse
further back, specifically $4^{(i-1)}$ lines above the error, with $i$ denoting
the current iteration. While this method does not guarantee error-free results,
it provides a more efficient and targeted strategy than simply reinitiating
sampling from the beginning.

%% file: sections/experiments.tex
\section{Experiments}\label{sec:experiments} 
Before fine-tuning our models on \dataset, we extract a sample of 1k
\emph{human-created} items to serve as our test set. As \llama's training
started in December 2022, we only sample items introduced after this date to
avoid data leakage. We conduct both automatic~(\secref{sec:auto-eval}) and
human evaluations~(\secref{sec:human-eval}). Additional results and instances
of generated \tikzname drawings are available in
\appappref{sec:additional-experiments}{sec:examples}.

\paragraph{Model Sizes}
In terms of model size, we fine-tune \llama[7b] and \clima[7b], each with 7
billion parameters~(7b), as well as \llama[13b] and \clima[13b] with 13 billion
parameters~(13b), respectively.  During inference, we additionally evaluate
\clima[13b] with \clip receiving compiled human-created \tikzname drawings as
input instead of captions, which we refer to as \clima[img] for clarity~(cf.\
\secref{sec:clima}).

\paragraph{Training}
Given the size of these models, we introduce trainable low-rank adaption
weights~(\lora; \citealp{hu2021lora}) while keeping the base model weights
frozen and in half precision~\citep{micikevicius2018mixed}. Following
\citet{dettmers2023qlora}, we apply \lora to all linear layers. In addition, we
find that training the embedding layer and language modeling heads is crucial
for successful fine-tuning. Since we are not aware of any studies applying
\lora to these layers, we make them fully trainable and leave this
investigation to future work.
In line with \citet{liu2023visual}, we train for 12 epochs with
\adam~\citep{loshchilov2019decoupled} and a batch size of 128, but increase the
learning rate to $5\mathrm{e}{-4}$ as this leads to faster convergence. As a
form of data augmentation only possible for \clima, we randomly replace the
captions forwarded to \clip with the reference image in 50\% of the cases.

\subsection{Automatic Evaluation}\label{sec:auto-eval}
We use a variety of automatic evaluation metrics to evaluate the performance of
our models on our test set in terms of code, image, and caption-image
similarity. In particular, we use the following metrics:

\begin{description}
  \item[\clipscore] calculates the similarity between image and text, as
    outlined in \secref{sec:data-augmentation}. We utilize it to evaluate the
    correlation between a rasterized \tikzname drawing and its corresponding
    caption.
  \item[{\clipscore[img]}] is technically the same metric as \clipscore, but
    with human-made \tikzname drawings as a reference input. Therefore, it
    assesses the similarity of two images rather than an image and a caption.
    To our best knowledge, we are the first to use \clipscore in this
    configuration.
  \item[Kernel Inception Distance~(\kid)] assesses the quality of generated
    \tikzname drawings by comparing their distribution with the distribution of
    real images in the test set~\citep{binkowski2018mmd}. This comparison helps
    determine how realistic the generated images appear in general. We extract
    image features using the same \clip model utilized in \clipscore.
  \item[\crystalbleu] is an n-gram-based metric designed to measure textual
    similarity~\citep{aryaz2023crystableu}. As a variant of
    \bleu~\citep{papineni2002bleu}, optimized for evaluating code, we employ it
    to assess the similarity between human-created and machine-produced
    \tikzname code.
  \item[Extended Edit Distance~(\eed)] is a metric dedicated to assessing
    string similarity~\citep{stanchev2019eed}, much like \crystalbleu. We
    utilize it to determine the minimum number of operations needed to convert
    the machine-generated code into the reference code.
  \item[Compilation Sampling Rate~(\csr)] measures how frequently we need to
    sample from a model to yield compilable \tikzname code that outputs an
    image. This is crucial as some metrics depend on images. With \llama and
    \clima, we use iterative resampling~(cf.\ \secref{sec:error-handling}) and
    account for partial samples. This is not feasible with \gpt and \claude due
    to their chain-of-thought prompt, which generates code across multiple
    steps. We take a relaxed stance, counting a sample as successful if it
    results in an image, even if there are errors.
\end{description}

\begin{figure}[tb]
  \centering
  \includegraphics{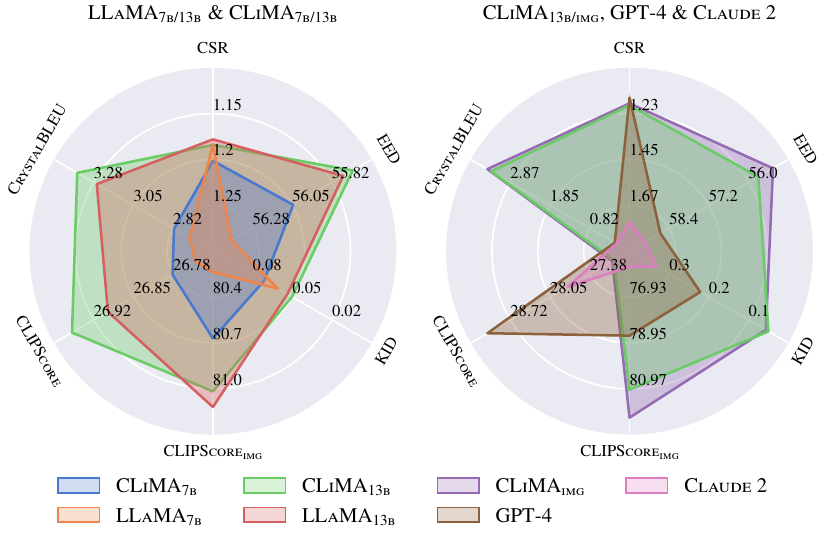}
  \caption{Automatic evaluation results for \llama[7b/13b], \clima[7b/13b/img],
  \gpt, and \claude. Axes representing metrics where lower values are
  better~(\csr, \eed, and \kid) have been inverted. Detailed scores are
  provided in \appref{sec:additional-experiments} for further reference.}%
  \label{fig:scores}
\end{figure}

\paragraph{Results}
We compute the above metrics for each model and present the system-level scores
in \figref{fig:scores}. The radar chart on the left illustrates that there are
small but noticeable score differences between the \llama and \clima models,
revealing some intriguing trends. Aligning with the intuitive expectation that
larger models yield better performance~\citep{kaplan2020scaling}, the 13b
models clearly outperform the 7b models on all string-similarity and
image-based metrics by 0.2--0.5\pp~(percentage points). A notable exception is
\csr, where all models perform comparably. This shows that all models require
approximately 1.2 samples per caption to generate a compilable \tikzname
drawing.

Within model sizes, \clima[7b] outperforms \llama[7b] on \crystalbleu, \eed,
\clipscore, and \clipscore[img] by up to 0.4\pp, suggesting that, even when
only text inputs are involved, integrating \clip into the model has a
predominantly positive effect.\ \clima[13b] continues this trend, showing a
0.1\pp higher \clipscore than \llama[13b]. However, we also see that this does
not necessarily have to increase the similarity with a reference image as well,
as \llama[13b] has a 0.1\pp higher \clipscore[img]. On \crystalbleu and \eed,
\clima[13b] again fares better, although, with 0.1\pp, the gap is not as
pronounced as for the 7b models, possibly due to diminishing
returns~\citep{hong2023diminishing}.

The right radar chart compares our best text-only model, \clima[13b], with
\clima[img], \gpt, and \claude. As before, all models perform roughly the same
on \csr, except for \claude, which needs noticeably more samples. As expected,
\clima[img], having access to reference images, improves upon \clima[13b] in
\clipscore[img] by 1.2\pp. However, this does not lead to an improvement in
\clipscore, echoing our earlier observation that image and caption-image
similarity do not always correlate. It also does not improve \kid,
demonstrating that the overall quality of the images remains constant.
Nevertheless, the string-based metrics are 0.1--0.4\pp higher, indicating that
conditioning on reference images positively impacts code similarity.

We also observe that \claude performs much worse than \gpt, and both perform
noticeably worse than both \clima[13b] and \clima[img] on most metrics. The
drastically lower \crystalbleu and \eed (up to 3.9\pp) suggest that \gpt and
\claude generate fundamentally different code~(in \appref{sec:complexity} we
show that it exhibits a lower level of complexity). The up to 6.6\pp lower
\clipscore[img] and over six times as large \kid indicate that not only do the
generated images look different from human ones, but also that the general
quality of images is much different from the human distribution. However, most
strikingly, both models achieve an up to 2.1\pp higher \clipscore. Upon
investigation, we find that both models tend to produce degenerate images,
which visibly copy the input caption into the output image. Since the outputs
of \clip~(and by extension \clipscore) can be controlled with \emph{images of
text}~\citep{ramesh2022dalle2}, \claude, and in particular \gpt, essentially
employ such typographic attacks to achieve exceptional caption-image
similarities. We further explore this phenomenon in \secref{sec:memorization}.

Overall, \clima[7b] and \clima[13b] outperform their respective \llama models
in five out of seven metrics each, with \claude and \gpt substantially
underperforming all of them. While \clima[img] unsurprisingly improves upon
\clima[13b], \clima[13b] is the best model with only textual inputs.

\subsection{Human Evaluation}\label{sec:human-eval}
\begin{figure}[tb]
  \centering
  \includegraphics{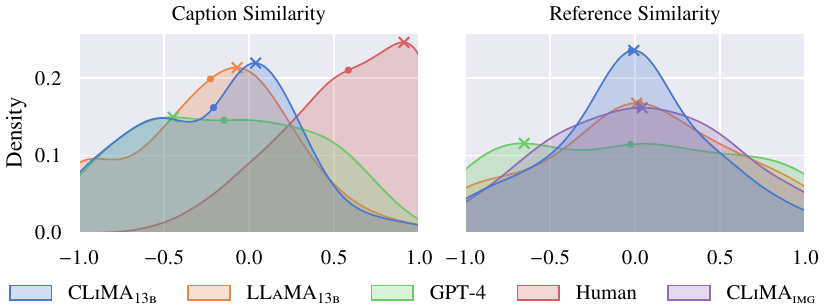}
  \caption{Distributions of \bws scores per model for caption and reference
  similarity. Scores span from -1~(poor) to 1~(excellent). The
  \enquote{\textbullet} markers denote expected values, and
  \enquote{\texttimes} signifies the mode.}%
  \label{fig:bws}
\end{figure}
To further evaluate the effectiveness of our models, we conduct a human
annotation campaign using \emph{best-worst scaling}~(\bws;
\citealp{louviere2015bws}). As a form of comparative annotation, \bws yields
high-quality results even when the number of annotations is
low~\citep{kiritchenko2016bws,kiritchenko2017bws}. Within this method,
annotators are tasked to compare tuples of $n=4$ items, identifying the best
and the worst item based on a given property. Real-valued scores, ranging from
-1~(poor) to 1~(excellent), are then computed by subtracting the fraction of
times an item is chosen as the worst from the fraction of times it is chosen as
the best~\citep{orme2009maxdiffa}.

In this work, we focus on \emph{caption similarity}~(\capsim) and
\emph{reference similarity}~(\refsim). In \capsim, annotators evaluate image
tuples based on text-image similarity with captions~(similar to \clipscore). We
construct 4-tuples consisting of our two leading text-only models from
automatic evaluation~(\clima[13b] and \llama[13b]), \gpt, and human reference
images. In \refsim, the human reference images are used as the standard of
comparison instead~(similar to \clipscore[img]), so we replace them in the
tuples with \clima[img], while leaving the other models unchanged.
Each property is then annotated by four unique expert annotators with relevant
research experience~(cf.\ \appref{sec:demographics}).\footnote{We tried
crowdsourcing as well, but due to low agreement with experts, we concluded that
crowdworkers lack the necessary expertise for our tasks~(cf.\
\appref{sec:demographics}).} To ensure a manageable workload for the
annotators, we create our tuples from a subset of 100 items sampled from our
test set. We assess the consistency of the annotators using \emph{split-half
reliability}~(\shr; \citealp{kiritchenko2017bws}). This method involves
randomly splitting all annotations into two sets, calculating scores for each
set, and then determining the correlation between them using Spearman's $\rho$.

\paragraph{Results}
For \capsim, the \shr is $\rho = 0.6$, indicating a moderate but adequate
consensus among annotators.\ \figref{fig:bws}~(left) exhibits kernel density
estimates for the computed scores, with marked modes and expected values.
Unsurprisingly, humans perform best with a mode near 1.\ \clima[13b] is the
only other model with a mode above 0, followed by \llama[13b], while \gpt lags
behind. This indicates that when sampling once with a given caption,
\clima[13b] is most likely to generate the best image. Since \clima[13b] and
\llama[13b] retain their earlier \clipscore ranking, but \gpt drops
substantially, we hypothesize that human annotators are not as prone to
typographic attacks as metrics. However, we still observe a slight bias towards
images of text. In 75\% of cases where \gpt is selected as the best model, it
copies more n-grams from the caption into the image than the worst-ranked
image, potentially leading to outliers and thus a slightly higher expected
value than \clima[13b] or \llama[13b].

Regarding \refsim, we record a similar \shr, with $\rho=0.58$. For \llama[13b],
\clima[13b], and \clima[img], the distributions in \figref{fig:bws}~(right) are
almost normally distributed, with the mode and expected value being nearly
identical. As with \clipscore[img], \llama[13b] is ranked marginally higher
than \clima[13b], indicating \clipscore[img] correlates well with human
rankings (\appref{sec:additional-experiments} details exact correlations). On a
similar scale, \clima[img] outperforms \llama[13b]. In contrast, \gpt follows a
nearly uniform distribution, with a slight downward trend for better scores.
Therefore, its mode is noticeably lower than for the other models. The expected
value, albeit only slightly, is also the lowest.

In summary, our human evaluation aligns well with our automatic metrics, with
the added benefit of lower susceptibility to typographic attacks.\ \clima[13b]
outperforms \llama[13b] on \capsim, while \clima[img] surpasses \llama[13b] on
\refsim. \gpt shows peculiar distributions, with the mode~(and also the
expected value for \refsim) lagging behind, highlighting the effectiveness of
our models.

%% file: sections/analysis.tex
\section{Analysis}\label{sec:memorization} 
The issue of language models memorizing and copying training data is a
prevalent
concern~\citep{mccoy2023raven,carlini2023memo,raunak2022memo,meehan2020sample}.
Similarly, we discovered in \secref{sec:auto-eval} that \gpt and \claude tend
to perform typographic attacks by memorizing and copying input captions. In
this section, we analyze the extent of these issues on our test set using the
concept of \emph{n-gram novelty}~\citep{mccoy2023raven}. Specifically, to
measure \emph{code novelty}, we determine the proportion of n-grams in the
model-generated \tikzname code that are \emph{not} found in the training data.
To measure \emph{caption copying}, we calculate the proportion of n-grams from
the caption that were copied verbatim into the output code. For comparison, we
also calculate both metrics for human references.

\begin{figure}[tb]
  \centering
  \includegraphics{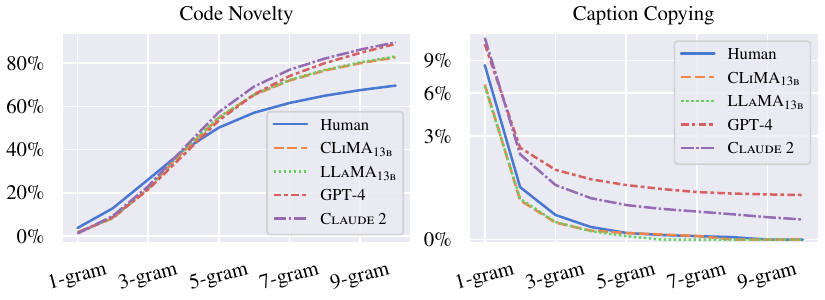}
  \caption{Proportion of unique code n-grams~($n \in [1,10]$) that do not
  appear in the training data~(left), and proportion of caption n-grams that
  were copied into the output image~(right).}%
  \label{fig:memorization}
\end{figure}
\figref{fig:memorization} displays the results for both metrics~($n \in
[1,10]$) after filtering code comments. In terms of code novelty, models tend
to generate less novel code than humans for smaller n-grams~($n<4$). However,
for $n>6$, models become more novel, with more than 80\% of all model n-grams
being novel for $n>8$.\ \citet{mccoy2023raven} observe the same phenomenon in
all datasets investigated and conclude that this ratio is the normal case when
a model is not affected by memorization of the training data.
Regarding caption copying, \gpt and \claude copy considerably more n-grams
from the caption than our models. For 1-grams~(i.e., $n=1$), \clima[13b] and
\llama[13b] copy around 6.5\% of n-grams, while \gpt and \claude copy more than
10\%. For $n > 5$, \clima[13b], \llama[13b], and humans practically stop
copying, but \claude and especially \gpt continue with an almost linear trend,
hinting at instances where they might copy the entire caption~(cf.\
\appref{sec:examples} for examples). This reinforces our hypothesis from
\secref{sec:auto-eval} and points towards \clipscore[img] as a more robust
metric for assessing the visuals of text-rich images since it seems less
susceptible to typographic attacks.

%% file: sections/conclusion.tex
\section{Conclusion \& Future Work} 
In this work, we present \projectname, a project for automatically generating
\tikzname drawings based on natural language descriptions. As part of
\projectname, we release \dataset, a pioneering dataset of aligned \tikzname
drawings and captions, and \clima, a novel model architecture which integrates
multimodal \clip embeddings into \llama. By fine-tuning \clima on \dataset, we
demonstrate that it outperforms \llama on several metrics and also surpasses
proprietary \gpt and \claude. In addition, \clima can also process images,
potentially extending its application to vectorization and conditioning on
sketches.
Important finds are that (i) integrating \clip can lead to improvements, even
when only text inputs are involved, provided the task relates to visual
concepts, and that (ii) attention should be paid to typographic attacks when
evaluating models that generate text-rich images.

In future research, we aim to incorporate insights from the caption generation
community and enrich our input texts with other figure-mentioning sections of
the source documents~(cf.\ \secref{sec:related-work}). We also plan to enhance
our extraction pipeline, especially since we had to exclude over 120k \tikzname
images from \arxiv that failed to compile. We hope that these modifications
will bring us a step closer to bridging the gap to human performance.

%% file: sections/ethics.tex
\section{Ethics Statement} 
We ensure that the \tikzname drawings we gather from online sources are
licensed in a manner that permits us to copy and redistribute them. Most
sources license their content under a Creative Commons attribution
license,\footnote{\url{https://creativecommons.org/licenses}} the GNU Free
Documentation
License,\footnote{\url{https://www.gnu.org/licenses/fdl-1.3.en.html}} or the
MIT license.\footnote{\url{https://opensource.org/license/mit}} \Arxiv is an
exception in that, even though it allows licensing under a Creative Commons
license, the majority of papers are published under a non-exclusive license,
which does not grant us permission to
redistribute.\footnote{\url{http://arxiv.org/licenses/nonexclusive-distrib/1.0}}
As a result, we exclude any \tikzname drawings from \arxiv that use this
license in the public release of \dataset. Nevertheless, we do release
\projectname in conjunction with the dataset generation code, enabling anyone
to recreate the full version of \dataset themselves. As for auto-generated
samples, \citeauthor{openai2023gpt4} prohibits the use of \gpt for creating
competing services, restricting this part of our dataset to non-commercial
applications.\footnote{\url{https://openai.com/policies/terms-of-use}}

Apart from our dataset, in this work we compare openly developed models with
the proprietary \gpt and \claude, whose full training details and
hyperparameters have not been published. While we strive for a fair evaluation
of all models, the lack of transparency in proprietary systems inevitably
hinders comparative evaluations and reproducibility.

Within the scope of these limiting aspects, however, our models perform best in
generating \tikzname across the tested conditions. Yet, our models should not
be used as a substitute for human judgment and critical thinking. They may
carry any biases, flaws, or gaps that exist in the base models and training
data and could potentially misinterpret the input, fabricate non-existent
details, or overlook crucial information. Users should be aware of potential
differences between the results they expect and the output the models generate.

Furthermore, while our models are designed to aid in the production of
legitimate scientific figures, they could potentially be used to generate
disinformation and fake science in the hands of malicious actors.

%% file: sections/acknowledgements.tex
\section*{Acknowledgments} 
We gratefully thank, in no particular order, Timm Dill, Yanran Chen, Daniil
Larionov, JiWoo Kim, Vivian Fresen, Martin Kerscher, Christoph Leiter, and Ran
Zhang for their help with our human evaluation campaign, proofreading,
discussions, and comments on our work. We further thank the BMBF for its
support via the grant \textsc{Metrics4NLG}\@. The second author is funded under
the Excellence Strategy of the German Federal Government and the Länder. The
last author is supported by DFG grant EG~375/5--1. We also thank \huggingface
for providing a community GPU grant. Any icons used in this work were designed
by \openmoji, the open-source emoji and icon project.

%% file: sections/appendix.tex
\section{Code \& Image Complexity}\label{sec:complexity} 
\begin{wraptable}[13]{R}{0pt}
  \centering
  \begin{tabular}[tb]{l d{3.2} *{2}{d{1.2}}}
    \toprule
    \thead{Source} & \nhead{Tokens} & \nhead{Cplx} & \nhead{Qlty}\\
    \midrule
    Human       & 916.56 & 0.76 & 0.83\\
    \clima[13b] & 422.56 & 0.46 & 0.4 \\
    \llama[13b] & 420.28 & 0.46 & 0.39\\
    \gpt        & 186.73 & 0.31 & 0.38\\
    \claude     & 179.42 & \nempty & \nempty\\
    \bottomrule
  \end{tabular}
  \caption{Average number of tokens in \tikzname documents, along with averaged
  min-max normalized \bws scores for image complexity~(Cmplx) and
  quality~(Qlty).}%
  \label{tab:complexity}
\end{wraptable}
An interesting phenomenon we observe in our experiments in
\secref{sec:experiments} is noticeable differences in \tikzname code size. As
shown in \tabref{tab:complexity}, human-created \tikzname drawings, after
filtering comments, contain twice the number of tokens as \clima[13b] and
\llama[13b], which in turn have twice as many tokens as \gpt and \claude. We
investigate how this discrepancy is reflected in the compiled images, in
particular, whether images created from code with fewer tokens are less complex
and whether this also affects image quality.
To this matter, we gather another round of \bws scores. Analogous to
\secref{sec:human-eval}, we form tuples with \clima[13b], \llama[13b], \gpt,
and humans. We assign a single annotator to evaluate image tuples in terms of
complexity and, due to the potentially subjective nature of the task, four
annotators to evaluate image quality. Since we obtain an \shr of $\rho=0.68$
for image quality, we conclude that all annotators have a similar idea of what
constitutes it.

We display averaged and min-max normalized \bws scores in
\tabref{tab:complexity}. Unsurprisingly, there is a clear correlation between
code size and image complexity. Humans score over twice as high as \gpt, with
\clima[13b] and \llama[13b] falling in between. A link to quality is also
visible, although less pronounced. Humans excel distinctly, but \clima[13b],
\llama[13b], and \gpt are not far apart. Nonetheless, in absolute terms,
\clima[13b] still ranks higher than \llama[13b], and \gpt comes in last.
Overall, this confirms our observations in \secref{sec:auto-eval} that \gpt and
\claude produce code most different from humans, yielding figures that are
simpler than other models. Consequently, we believe that our tool can assist
humans better in creating more detailed \tikzname drawings.

\section{Prompt Engineering}\label{sec:prompting} 
Our work involves developing a series of prompts to guide general-purpose
language models in accomplishing specific tasks. In this section, we discuss
each model and the corresponding prompts we designed. Within a prompt, terms
enclosed in curly braces symbolize placeholder variables that are substituted
during inference.

\paragraph{\wizardlm}
We employ \wizardlm to create captions for \tex Stack Exchange content based
on a question and its title~(cf.\ \secref{sec:data-acquisition}). We force the
generated tokens to start with \enquote{\texttt{Desired outcome:}} which we
subsequently strip from the output, along with any text generated after the
first newline:
\begin{prompt}
  Create a clear and specific caption for an image that depicts the desired
  outcome of the following question. Utilize all relevant details provided in
  the question, particularly focusing on the visual aspects. Avoid referencing
  any example images or code snippets. Ensure that your caption is
  comprehensive and accurate:\\\\\{TITLE\}\\\\\{QUESTION\}
\end{prompt}

\paragraph{\llavar}
We use \llavar to generate short descriptions for scientific figures~(cf.\
\secref{sec:data-augmentation}). We initially tried to provide \llavar with the
original caption as context to improve it directly, but in many cases this
seemed to confuse the model. Therefore, we decided to create short descriptions
based solely on the figure and later append them to the original caption. We
found that short, concise prompts work best for generating these descriptions:
\begin{prompt}
  \{IMAGE\}\\Write a short description for the image.
\end{prompt}

\paragraph{\gpt \& \claude}
To generate \tikzname drawings~(cf.\ \secref{sec:data-acquisition},
\secref{sec:methods}, and \secref{sec:experiments}), we use the same prompt for
\gpt and \claude, which was originally designed by
\citet{zhang2023controllable} and slightly modified for our task. After a
comparison with our own prompts and prompts from other
works~\citep{bubeck2023sparks}, we found it to work best in our initial
experiments. As a chain-of-thought prompt it breaks down the task of code
generation into a series of logical steps. By default, the models tend to copy
the input caption into the output image. To counter this, we additionally
instruct the models not to include the caption in generated images. However,
instances of caption copying still occur, as discussed in
\secref{sec:memorization}:
\begin{prompt}
  Draw a TikZ picture for the following caption: \{CAPTION\}. First, you need
  to provide an abstract step-by-step drawing guide. Then, you need to generate
  the full code~(beginning with
  \textquotedbl\textbackslash{}documentclass\textquotedbl\ and ending with
  \textquotedbl\textbackslash{}end\{document\}\textquotedbl) following the
  guide. Avoid any direct inclusion of the caption or any lengthy text
  sequences within the code. Finally, summarize the drawing.
\end{prompt}

\section{Additional Details on Experiments}\label{sec:additional-experiments}
In this section, we conduct additional ablation studies and provide more
specific aspects of our experiments in \secref{sec:experiments}. This will
provide a clearer understanding of our methodologies and their respective
considerations and limitations.

\subsection{Training Challenges}
During the training phase, we limit the context size of our models to 1200
tokens due to constraints of our existing GPU resources, and filter out all
examples that exceed this limit. Although most examples were within these
limits, they were generally very close to the maximum allowable length.
This prevented us from training our models to perform functions such as
multi-turn conversations like chatbots, which could have allowed compiler logs
to be fed back into the model to obtain improved outputs. Although we did
consider this approach in preliminary experiments with vanilla \llama and its
instruction-tuned derivatives in a zero-shot setting, the baseline performance
of these models was too poor to produce usable \tikzname output.
Instead, by drawing parallels to text-to-image generation models that condition
the output solely on the input caption and using our iterative resampling
technique to improve the outputs, we provide more resource-efficient
alternatives with good performance.

\subsection{In-depth Evaluation Metric Scores}
\tabref{tab:scores} presents the precise system-level scores from our automatic
evaluation in \secref{sec:auto-eval}. As observed in that section, \clima[img]
and \clima[13b] generally outperform the other models, with \clipscore being
a notable exception due to \gpt and \claude being prone to caption copying.
\begin{table}
  \centering
  \newcommand{\first}[2]{\multicolumn{1}{B{#1.3}}{#2}}
  \newcommand{\second}[2]{\multicolumn{1}{U{#1.3}{#2}}{#2}}
  \newcommand{\down}{\unskip\rlap{\textsubscript{\textdownarrow}}}
  \newcommand{\up}{\unskip\rlap{\textsubscript{\textuparrow}}}
  \begin{tabular}{l *{2}{d{1.3}} d{2.3} d{1.3} *{2}{d{2.3}} d{1.3} }
    \toprule
    \thead{Model} & \nhead{\cer\down} & \nhead{\csr\down} & \nhead{\eed\down} & \nhead{\kid\down} &
    \nhead{\clip[img]\up} & \nhead{\clip\up} & \nhead{\bleu\up}\\
    \midrule
    \llama[7b]  & 0.644 & 1.183 & 56.393 & 0.059 & 80.237 & 26.733 & 2.732\\
    \llama[13b] & 0.552 & 1.178 & 55.762 & \second{1}{0.053} & 81.12  & 26.898 & 3.258\\
    \midrule
    \clima[7b]  & \second{1}{0.545} & 1.202 & 56.045 & 0.068 & 80.671 & 26.776 & 2.82 \\
    \clima[13b] & 0.656 & 1.184 & \second{2}{55.708} & \first{1}{0.05}  & \second{2}{81.017} & 26.966 & \second{1}{3.369}\\
    \clima[img] & 0.658 & \second{1}{1.175} & \first{2}{55.264} & 0.057 & \first{2}{82.252} & 26.994 & \first{1}{3.469}\\
    \midrule
    \claude     & 1.115 & 1.758 & 59.135 & 0.333 & 75.589 & \second{2}{27.753} & 0.137\\
    \gpt        & \first{1}{0.384} & \first{1}{1.147} & 58.667 & 0.222 & 78.63  & \first{2}{29.115} & 0.181\\
    \bottomrule
  \end{tabular}
  \caption{Precise system-level scores for automatic evaluation metrics.
  \clipscore is abbreviated with \clip and \crystalbleu with \bleu. \cer shows
  the average number of compile time errors. Arrows indicate whether larger or
  smaller scores are better, and the best scores are visually highlighted.}%
  \label{tab:scores}
\end{table}
The table also contains the Code Error Rate~(\cer), which indicates the average
number of compile time errors of our models. As evidenced, it largely reflects
the trends of \csr. Most models generate less than one error on average, with
\claude being the only exception. Overall, \gpt achieves the best score,
possibly because it has fewer opportunities to make mistakes by generating
shorter and less complex code than other models~(cf.\ \secref{sec:complexity}).
Since \clipscore, at a fundamental level, evaluates similar aspects to \capsim
from our human evaluation in \secref{sec:human-eval}~(caption-image
similarity), and \clipscore[img] is the intuitive counterpart to \refsim~(image
similarity), we also calculate their respective Spearman correlations. At the
segment-level, the correlation coefficients are 0.23 for \clipscore and
\capsim, and 0.25 for \clipscore[img] and \refsim~(the degrees of correlation
aligning with values commonly found for machine translation metrics;
\citealp{freitag2021experts,rei2022comethino}). The system-level correlations
are 0.5 and 0.8, respectively. The higher correlations witnessed for image over
caption-image similarity lends credibility to our hypothesis regarding
\clipscore's biases favoring images of text.

\subsection{Ablation Studies}
In this subsection we delve into ablation studies performed to understand the
degree to which data augmentation and the different data sources of \dataset
contribute to the test set performance of our models.

\paragraph{Impact of Data Augmentation}
In our work, we incorporate two key data augmentation techniques: we~(i)
automatically augment image captions in \dataset~(cf.\
\secref{sec:data-augmentation}), and we~(ii) randomly replace image captions
forwarded to \clip with reference images during the training of \clima~(cf.\
\secref{sec:experiments}). Even though we show that caption augmentation
improves caption-image alignment, we have yet to conclusively quantify its
effect on model performance. Similarly, although mixing text-only and
text-image modalities is common when training models accepting both unimodal
and multimodal input~\citep{li2020unicoder,wang2022ofa}, the implications of
our particular training strategy remain to be explicitly measured. To
investigate these matters, we conduct an ablation study using \clima[7b] as a
representative model for efficiency reasons. In particular, we construct a
version of \clima[7b] that uses \dataset without caption augmentation and
another version where we train without sampling images. The performance on our
evaluation metrics compared to full training results of \clima[7b] can be seen
in \tabref{tab:ablation}.
\begin{table}
  \centering
  \newcommand{\down}{\unskip\rlap{\textsubscript{\textdownarrow}}}
  \newcommand{\up}{\unskip\rlap{\textsubscript{\textuparrow}}}
  \newcommand{\minus}[1]{$-\textrm{#1}$}

  \newcommand{\gradient}[5][0]{
    \pgfmathparse{(#1 ? (#3-#2) : (#2-#3))<0} 
    \xdef\isworse{\pgfmathresult}
    \xdef\bgclr{\ifnum\isworse=1 worse\else better\fi}

    \pgfmathparse{int(round(100*pow(abs((#2-#3)/(\isworse ? #5-#3 : #4-#3)), 1/3)))} 
    \xdef\tempa{\pgfmathresult}

    \cellcolor{\bgclr!\tempa} #2
  }

  \newcommand{\cergrd}[1]{\gradient[1]{#1}{0.545}{0.503}{2.145}}
  \newcommand{\csrgrd}[1]{\gradient[1]{#1}{1.202}{1.169}{1.546}}
  \newcommand{\eedgrd}[1]{\gradient[1]{#1}{56.045}{56.246}{60.552}}
  \newcommand{\kidgrd}[1]{\gradient[1]{#1}{0.068}{0.056}{0.229}}
  \newcommand{\imggrd}[1]{\gradient{#1}{80.671}{80.461}{75.15}}
  \newcommand{\clipgrd}[1]{\gradient{#1}{26.776}{26.958}{25.328}}
  \newcommand{\bleugrd}[1]{\gradient{#1}{2.82}{2.936}{0.886}}

  \setlength{\extrarowheight}{\belowrulesep}
  \setlength{\belowrulesep}{0pt}

  \begin{tabular}{l *{2}{d{1.3}} d{2.3} d{1.3} *{2}{d{2.3}} d{1.3} }
    \toprule
    \thead{Training Configuration} & \nhead{\cer\down} & \nhead{\csr\down} & \nhead{\eed\down} &
    \nhead{\kid\down} & \nhead{\clip[img]\up} & \nhead{\clip\up} & \nhead{\bleu\up}\\
    \midrule
    Full Training  & 0.545 & 1.202 & 56.045 & 0.068 & 80.671 & 26.776 & 2.82 \\
    \midrule
    \minus{Caption Augmentation} & \cergrd{0.603} & \csrgrd{1.169} & \eedgrd{56.384} & \kidgrd{0.056} & \imggrd{78.312} & \clipgrd{25.703} & \bleugrd{2.714}\\
    \minus{Image Training}       & \cergrd{0.681} & \csrgrd{1.178} & \eedgrd{56.336} & \kidgrd{0.08}  & \imggrd{80.129} & \clipgrd{26.604} & \bleugrd{2.736}\\[\dimexpr-\aboverulesep-.5\cmidrulewidth]
    \cmidrule(lr){1-1}\\[\dimexpr-\normalbaselineskip-\extrarowheight-.5\cmidrulewidth]
    \minus{\Arxiv Papers (86k)}       & \cergrd{2.145} & \csrgrd{1.546} & \eedgrd{60.552} & \kidgrd{0.229} & \imggrd{75.15}  & \clipgrd{25.328} & \bleugrd{0.886}\\
    \minus{\tex Stack Exchange (29k)} & \cergrd{0.624} & \csrgrd{1.207} & \eedgrd{56.274} & \kidgrd{0.071} & \imggrd{80.461} & \clipgrd{26.958} & \bleugrd{2.936}\\
    \minus{Artificial Examples (4k)}  & \cergrd{0.503} & \csrgrd{1.194} & \eedgrd{56.51}  & \kidgrd{0.079} & \imggrd{79.912} & \clipgrd{26.68}  & \bleugrd{2.537}\\
    \rule[-\dp\strutbox-\aboverulesep]{0pt}{0pt}%
    \minus{Curated Examples (1k)}     & \cergrd{0.656} & \csrgrd{1.22}  & \eedgrd{56.246} & \kidgrd{0.072} & \imggrd{80.346} & \clipgrd{26.815} & \bleugrd{2.711}\\[-\aboverulesep]
    \bottomrule
  \end{tabular}
  \caption{Ablation study results for \clima[7b] demonstrating how omitting
  either caption augmentation or randomly forwarding images to \clip instead of
  captions during training impacts performance on the test set. The study also
  evaluates the impact of each data source by training \clima[7b] on subsets of
  \dataset, omitting one data source each time. Scores lower or higher than
  those from full training~(taken from \tabref{tab:scores}) are shaded in red
  and green, respectively. Color intensity mirrors each score's proportion to
  the minimum and maximum values of its metric.}%
  \label{tab:ablation}
\end{table}

Although eliminating caption augmentation reveals subtle improvements for \csr
and \kid, the model noticeably underperforms for all other metrics,
particularly \clipscore~(1.1pp lower) and \clipscore[img]~(2.4pp lower).
Interestingly, the metrics that show an improvement do not compare model
outputs with individual references but either do not use references at
all~(\csr) or use them only to evaluate general characteristics~(\kid). On the
other hand, metrics with worse results are mostly reference-based. We attribute
this observation to the reduced amount of information carried by non-augmented
captions, most likely leading to a decline in \clima's ability to capture the
connection between captions and images and, therefore, delivering lower
performance in reference-based metrics. Conversely, with less information that
might otherwise constrain the output space, the model is likely to gain more
flexibility in producing figures. We conjecture that this allows the model to
concentrate more on aspects that contribute positively to metrics without
direct references, such as \csr and \kid. This hypothesis provides a plausible
explanation for our observed results.

Upon examining the impacts of not sampling images during the training process,
we find that most metrics exhibit worse performance, with only \csr yielding
comparable results. These findings support our choice of training strategy and
show that integrating images throughout the training process can bolster
performance even when only textual inputs are considered at the evaluation
stage.

In summary, we observe that both data augmentation techniques contribute
positively to model performance. Caption augmentation enhances the
caption-image alignment, benefiting \clima[7b], while leveraging images during
training results in improvements as well.

\paragraph{Importance of each Data Source}
Given the diverse data sources making up \dataset~(\arxiv, \tex Stack Exchange,
curated examples, and artificial examples; cf.\ \secref{sec:data-acquisition}),
we aim to assess the contribution of each source to the test set performance.
This exploration serves multiple purposes: firstly, it enables us to justify
the integration of each source; secondly, it can inform future data collection
initiatives. To realize this objective, we develop four additional variants of
\clima[7b], where each variant is trained by omitting one data source and
utilizing only the remaining three. As before, we plot the difference in
performance compared to the full training results in \tabref{tab:ablation},
which should expose the importance of each data source.

The greatest performance decline occurs with the removal of \arxiv, a
consistent outcome across all metrics. As the primary contributor of scientific
figures totaling 86k, its absence instigates a decline of 4.5pp for \eed, 5.5pp
for \clipscore[img], 1.4pp for \clipscore, and 1.9pp for \crystalbleu.
Similarly, \cer inflates nearly fourfold, and \csr experiences a noticeable
increase, as well.
Strikingly, however, results diverge for \dataset's second largest source, the
\tex Stack Exchange, which provides 29k examples. Contrary to expectations, the
general downward trend is comparatively small, with \eed and \clipscore[img]
only suffering a 0.2pp decrease and \clipscore and \crystalbleu even improving
by 0.2pp and 0.1pp, respectively. We attribute this outcome to \tex Stack
Exchange's unique focus on providing problem-specific minimum working examples
rather than scientific figures like the rest of \dataset. Although this could
still provide valuable training signals, we did not design our test set~(only
\textasciitilde100 of its 1k random samples originate from this source) to
evaluate the ability of our models to follow technical advice given in the
figure captions. Consequently, we plan to improve our testing framework in
future work.
Next, although we exclude artificial examples from our test set due to their
non-human origin and their removal from the training data improves \cer and
\csr, their absence still has a negative impact on all other metrics. Without
artificial examples, \eed drops by 0.5pp, \clipscore[img] by 0.8pp, \clipscore
by 0.1pp and \crystalbleu by 0.3pp, showing that distillation of \gpt is a
useful component in the training process.
Lastly, even though curated examples constitute the smallest data source~(less
than 1k examples), their omission from the training data induces greater negative
performance impacts than the removal of \tex Stack Exchange (0.2pp
for \eed, 0.3pp for \clipscore[img] and 0.1pp for \crystalbleu). We attribute
this disproportionate value per size to the likely higher quality of code and
captions inherent to curated examples.

Overall, if the results are examined independently, we observe that each
source has primarily positive effects. However, when evaluated in the
context of the source sizes, we find that curated examples create a
particularly noticeably positive impact, whereas \tex Stack Exchange exhibits
smaller effects. These insights will guide us to focus more on collecting such
curated examples in future work.

\section{Annotator Demographics}\label{sec:demographics}
Our annotators are proficient in English, possessing a C1 level or above as per
the Common European Framework of Reference for
Languages.\footnote{\url{https://www.coe.int/lang-cefr}} They all come from a
science and technology background with research experience. They specifically
consist of one female and one male faculty member, one female PhD student,
three male PhD students, and two female and two male assistants from other
institutions.

In addition to these expert annotators, we also conducted preliminary tests
with crowdworkers via Amazon Mechanical
Turk,\footnote{\url{https://www.mturk.com}} gathering ten crowd-annotations per
tuple for each task. However, the correlation between the results of the
crowdworkers and the experts was strikingly low~($\rho < 0.1$). Since the
crowdworkers also showed low agreement among themselves, we decided to
discontinue further experiments with crowdsourcing.

\section{Examples}\label{sec:examples}
In \figref{fig:dataset-examples}, we provide human-created examples from
\dataset. Further, \figref{fig:examples} contains examples of scientific
figures generated by \clima[13b], \llama[13b], and \gpt for our human
evaluation~(cf.\ \secref{sec:human-eval}). Each model is represented by one
high-scoring and one low-scoring instance, as assessed by our expert annotators
in terms of caption similarity. It is worth noting that the low-scoring example
of \gpt suffers from caption copying. We provide additional examples of
caption copying in \figref{fig:caption-copying}. Instances of generated code
can be found in \figref{fig:example-code}.
\begingroup
\usetikzlibrary{arrows} 
\newlength{\figheight}
\newlength{\capwidth}
\newcommand{\good}{\textcolor{OliveGreen}{good}}
\newcommand{\bad}{\textcolor{BrickRed}{bad}}
\newcommand{\inputcap}[2]{\def\getCaption{}\input{examples/#2/#1.tex}}
\newcommand{\inputimg}[2]{\resizebox{!}{\figheight}{\input{examples/#2/#1.tex}}}
\newcommand{\modelcap}[3]{%
  \renewcommand\thesubfigure{#1~(#2)}
  \captionsetup{font=scriptsize,labelformat=simple,labelsep=colon}%
  \caption{#3}%
}
\newtcolorbox{example}[1][]{
  enhanced jigsaw,
  colback=examplebg,
  colframe=exampletitle,
  halign=center,
  valign=center,
  boxsep=-1pt,
  boxrule=0pt,
  left=2mm,
  right=2mm,
  #1
}
\newtcolorbox{titleexample}[2][]{
  enhanced jigsaw,
  colback=examplebg,
  colframe=exampletitle,
  coltitle=black,
  halign=center,
  valign=center,
  fonttitle=\small,
  center title,
  boxsep=-1pt,
  boxrule=0pt,
  left=2mm,
  right=2mm,
  bottom=2mm-\tcboxedtitleheight,
  attach boxed title to bottom left={xshift=0cm,yshift*=\tcboxedtitleheight},
  boxed title style={left=0.25mm, right=0.25mm, boxsep=1.4mm, boxrule=0pt,
    sharp corners=downhill, colback=exampletitle},
  title={#2},
  #1
}
\newtcbinputlisting{\tikzlisting}[2][]{%
  enhanced jigsaw,
  colback=examplebg,
  colframe=exampletitle,
  boxsep=-1pt,
  boxrule=0pt,
  left=2mm,
  right=2mm,
  listing only,
  minted language=latex,
  minted options={numbersep=\dimexpr12pt-2mm,breaklines=true,breaksymbol={},linenos=true},
  listing file={#2},
  #1
}
\newcommand{\modelrow}[2]{%
  \begin{subfigure}[t]{\capwidth}
    \begin{example}
      \resizebox{!}{\figheight}{\input{examples/good/#1.tex}}
    \end{example}
    \modelcap{#2}{\good}{\def\getCaption{}\input{examples/good/#1.tex}}
  \end{subfigure}\hfill%
  \begin{subfigure}[t]{\capwidth}
    \begin{example}
      \resizebox{!}{\figheight}{\input{examples/bad/#1.tex}}
    \end{example}
    \modelcap{#2}{\bad}{\def\getCaption{}\input{examples/bad/#1.tex}}
  \end{subfigure}
}

\begin{figure}
  \setlength{\figheight}{3cm}
  \setlength{\capwidth}{.49\linewidth}

  \begin{subfigure}[t]{\capwidth}
    \begin{example}
      \resizebox{!}{\figheight}{\input{examples/datikz/1.tex}}
    \end{example}
    \caption{\def\getCaption{}\input{examples/datikz/1.tex}}
  \end{subfigure}\hfill%
  \begin{subfigure}[t]{\capwidth}
    \begin{example}
      \resizebox{!}{\figheight}{\input{examples/datikz/2.tex}}
    \end{example}
    \caption{\def\getCaption{}\input{examples/datikz/2.tex}}
  \end{subfigure}\medskip

  \setlength{\figheight}{4cm}
  \begin{subfigure}[t]{\capwidth}
    \begin{example}
      \resizebox{!}{\figheight}{\input{examples/datikz/3.tex}}
    \end{example}
    \caption{\def\getCaption{}\input{examples/datikz/3.tex}}
  \end{subfigure}\hfill%
  \begin{subfigure}[t]{\capwidth}
    \begin{example}
      \resizebox{!}{\figheight}{\input{examples/datikz/4.tex}}
    \end{example}
    \caption{\def\getCaption{}\input{examples/datikz/4.tex}}
  \end{subfigure}
  \caption{Illustrative examples from the \dataset, licensed for free
    distribution. The examples are from \url{https://github.com/PetarV-/TikZ},
    \url{https://github.com/janosh/tikz}, \url{https://tikz.net}, and
    \url{https://arxiv.org}, respectively.}%
  \label{fig:dataset-examples}
\end{figure}
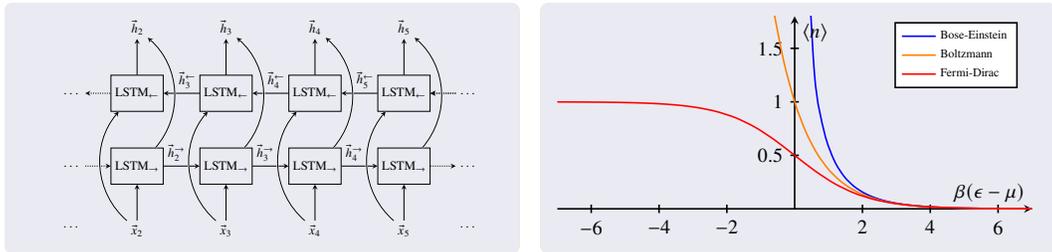
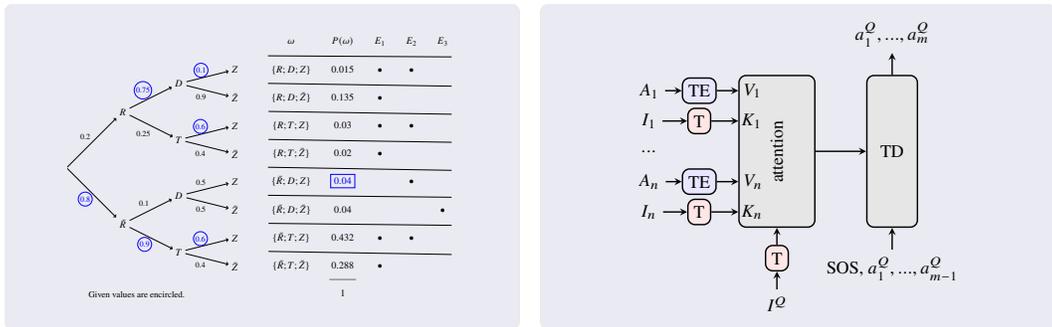

\begin{figure}
  \centering
  \setlength{\figheight}{4cm}
  \setlength{\capwidth}{.49\linewidth}

  \modelrow{clima-13b}{\clima[13b]}
  \par\medskip\modelrow{llama-13b}{\llama[13b]}
  \par\medskip\modelrow{gpt-4}{\gpt}

  \caption{Examples of model-generated scientific figures that received
  high ratings~(\good) and low ratings~(\bad) according to the perception of
  our expert annotators for caption similarity. The sections of the captions
  that have been augmented~(cf.\ \secref{sec:data-augmentation}) are
  \emph{emphasized}.}%
  \label{fig:examples}
\end{figure}

\begin{figure}
  \setlength{\figheight}{3.5cm}

  \begin{subfigure}[t]{\textwidth}
    \begin{titleexample}{\gpt}
      \resizebox{!}{\figheight}{\input{examples/copying/gpt-4/1.tex}}
    \end{titleexample}
    \begin{titleexample}{\clima[13b]}
      \resizebox{!}{\figheight}{\input{examples/copying/clima-13b/1.tex}}
    \end{titleexample}
    \caption{\def\getCaption{}\input{examples/copying/gpt-4/1.tex}}%
  \end{subfigure}\medskip

  \setlength{\figheight}{4cm}
  \begin{subfigure}[t]{\textwidth}
    \begin{tcbraster}[raster columns=2,raster equal height]
      \begin{titleexample}{\gpt}
        \resizebox{!}{\figheight}{\input{examples/copying/gpt-4/2.tex}}
      \end{titleexample}
      \begin{titleexample}{\clima[13b]}
        \resizebox{!}{\figheight}{\input{examples/copying/clima-13b/2.tex}}
      \end{titleexample}
    \end{tcbraster}
    \caption{\def\getCaption{}\input{examples/copying/gpt-4/2.tex}}%
  \end{subfigure}\medskip

  \begin{subfigure}[t]{\textwidth}
    \begin{tcbraster}[raster columns=2,raster equal height]
      \begin{titleexample}{\gpt}
        \quad\resizebox{!}{\figheight}{\input{examples/copying/gpt-4/3.tex}}
      \end{titleexample}
      \begin{titleexample}{\clima[13b]}
        \quad\resizebox{!}{\figheight}{\input{examples/copying/clima-13b/3.tex}}
      \end{titleexample}
    \end{tcbraster}
    \caption{\def\getCaption{}\input{examples/copying/gpt-4/3.tex}}%
    \label{fig:caption-copying-3}
  \end{subfigure}

  \caption{Additional examples of \gpt that suffer from caption copying. For
  reference, we also provide the output of \clima[13b] for the same captions.
  The underlying \tikzname code of \figref{fig:caption-copying-3} can be found
  in \figref{fig:example-code}.}%
  \label{fig:caption-copying}
\end{figure}

\begin{figure}
  \small
  \begin{subfigure}[t]{\textwidth}
    \tikzlisting{figures/examples/copying/gpt-4/3-full.tex}
    \renewcommand\thesubfigure{\gpt\unskip}
    \captionsetup{labelformat=simple,labelsep=colon}%
    \caption{\def\getCaption{}\input{examples/copying/gpt-4/3.tex}}%
  \end{subfigure}\medskip

  \begin{subfigure}[t]{\textwidth}
    \tikzlisting{figures/examples/copying/clima-13b/3-full.tex}
    \renewcommand\thesubfigure{\clima[13b]\unskip}
    \captionsetup{labelformat=simple,labelsep=colon}%
    \caption{\def\getCaption{}\input{examples/copying/clima-13b/3.tex}}%
  \end{subfigure}
  \caption{The generated code of \gpt~(top) and \clima[13b]~(bottom) from
    \figref{fig:caption-copying-3}. Both models correctly use the
    \texttt{overlay} option, but they are not devoid of issues. In particular,
    \gpt utilizes nested \texttt{tikzpicture} environments which is generally
    discouraged and the output of \clima[13b] seems particularly verbose.}%
  \label{fig:example-code}
\end{figure}
\endgroup

%% file: figures/examples/datikz/1.tex
\ifdefined\getCaption%
A diagram representing a recurrent neural network consisting of several LSTM
blocks, processing the input sequence simultaneously forwards and backwards~(to
exploit both directions of temporal dependence). Contains some rather tight
manoeuvering.
\else
\usetikzlibrary{positioning}
\begin{tikzpicture}
        \node[rectangle] (Y0) at (0, 0) {$\dots$};
        \node[rectangle, draw, right=2em of Y0, minimum height=1cm, minimum width=1cm] (RNN) {LSTM$_\rightarrow$};
        \node[rectangle, right=of RNN, draw, minimum height=1cm, minimum width=1cm] (RNN2) {LSTM$_\rightarrow$};
        \node[rectangle, right=of RNN2, draw, minimum height=1cm, minimum width=1cm] (RNN3) {LSTM$_\rightarrow$};

        \node[rectangle, right= of RNN3, draw, minimum height=1cm, minimum width=1cm] (RNN4) {LSTM$_\rightarrow$};
        \node[rectangle, right=2em of RNN4] (RNN5) {$\dots$};

        \node[rectangle, above=of RNN4, draw, minimum height=1cm, minimum width=1cm] (R25) {LSTM$_\leftarrow$};
        \node[rectangle, left=of R25, minimum height=1cm, minimum width=1cm, draw] (R24) {LSTM$_\leftarrow$};
        \node[rectangle, left=of R24, draw, minimum height=1cm, minimum width=1cm] (R23) {LSTM$_\leftarrow$};
        \node[rectangle, left=of R23, draw, minimum height=1cm, minimum width=1cm] (R22) {LSTM$_\leftarrow$};
        \node[rectangle, left=2em of R22] (R21) {$\dots$};
        \node[right=2em of R25] (Y20) {$\dots$};

        \node[below=of RNN] (X1) {$\vec{x}_2$};
        \node[below=of RNN2] (X2) {$\vec{x}_3$};
        \node[below=of RNN3] (X3) {$\vec{x}_4$};
        \node[below=of RNN4] (X4) {$\vec{x}_5$};
        \node[above=of R25] (Y5) {$\vec{h}_5$};
        \node[above=of R24] (Y4) {$\vec{h}_4$};
        \node[above=of R23] (Y3) {$\vec{h}_3$};
        \node[above=of R22] (Y2) {$\vec{h}_2$};

        \draw[-stealth, thick] (X1) -- (RNN);
        \draw[-stealth, thick] (X2) -- (RNN2);
        \draw[-stealth, thick] (X3) -- (RNN3);
        \draw[-stealth, thick] (X4) -- (RNN4);
        \draw[-stealth, thick, densely dotted] (Y0) -- (RNN);
        \draw[-stealth, thick] (RNN) -- node[above, pos=0.35] {$\vec{h}_2^\rightarrow$} (RNN2);
        \draw[-stealth, thick] (RNN2) -- node[above, pos=0.35] {$\vec{h}_3^\rightarrow$} (RNN3);
        \draw[-stealth, thick] (RNN3) -- node[above, pos=0.35] {$\vec{h}_4^\rightarrow$} (RNN4);
        \draw[-stealth, densely dotted, thick] (RNN4) -- (RNN5);
        \node[below=4em of Y0] (d) {\dots};
        \node[below=4em of RNN5] (d) {\dots};

        \path[-stealth, ultra thick, white] (X1) edge[bend left=45] (R22);
        \path[-stealth, thick] (X1) edge[bend left=45] (R22);
        \path[-stealth, ultra thick, white] (X2) edge[bend left=45] (R23);
        \path[-stealth, thick] (X2) edge[bend left=45] (R23);
        \path[-stealth, ultra thick, white] (X3) edge[bend left=45] (R24);
        \path[-stealth, thick] (X3) edge[bend left=45] (R24);
        \path[-stealth, ultra thick, white] (X4) edge[bend left=45] (R25);
        \path[-stealth, thick] (X4) edge[bend left=45] (R25);
        \draw[-stealth, densely dotted, thick] (Y20) -- (R25);

        \draw[-stealth, thick] (R22) -- (Y2);
        \draw[-stealth, thick] (R23) -- (Y3);
        \draw[-stealth, thick] (R24) -- (Y4);
        \draw[-stealth, thick] (R25) -- (Y5);

        \draw[stealth-, densely dotted, thick] (R21) -- (R22);
        \draw[stealth-, thick] (R22) -- node[above, pos=0.65] {$\vec{h}_3^\leftarrow$} (R23);
        \draw[stealth-, thick] (R23) -- node[above, pos=0.65] {$\vec{h}_4^\leftarrow$} (R24);
        \draw[stealth-, thick] (R24) -- node[above, pos=0.65] {$\vec{h}_5^\leftarrow$} (R25);
        \draw[-stealth, densely dotted, thick] (Y20) -- (R25);

        \path[-stealth, ultra thick, white] (RNN) edge[bend right=45] (Y2);
        \path[-stealth, thick] (RNN) edge[bend right=45] (Y2);
        \path[-stealth, ultra thick, white] (RNN2) edge[bend right=45] (Y3);
        \path[-stealth, thick] (RNN2) edge[bend right=45] (Y3);
        \path[-stealth, ultra thick, white] (RNN3) edge[bend right=45] (Y4);
        \path[-stealth, thick] (RNN3) edge[bend right=45] (Y4);
        \path[-stealth, ultra thick, white] (RNN4) edge[bend right=45] (Y5);
        \path[-stealth, thick] (RNN4) edge[bend right=45] (Y5);

\end{tikzpicture}
\fi

%% file: figures/examples/datikz/2.tex
\ifdefined\getCaption%
A plot comparing the distribution functions of Bose-Einstein, Boltzmann, and
Fermi-Dirac statistics as a function of the reduced chemical potential $\beta
(\epsilon - \mu)$. This visualiation highlights the differences between the
three types of distribution functions, which are used to describe the behavior
of particles in different statistical systems.
\else
\begin{tikzpicture}
  \begin{axis}[
      xlabel = $\beta (\epsilon - \mu)$,
      ylabel = $\langle n\rangle$,
      ymin = 0,ymax = 1.8,
      smooth,thick,
      axis lines = center,
      every tick/.style = {thick},
      legend cell align=left,
      legend style={legend pos=north east,font=\tiny},
      width=10cm,height=5cm]

    \def\xmax{7}
    \addplot[color=blue,domain=0:\xmax]{1/(e^x - 1)};
    \addplot[color=orange,domain=-1:\xmax]{1/e^x};
    \addplot[color=red,domain = -\xmax:\xmax]{1/(e^x + 1)};

    \legend{Bose-Einstein,Boltzmann,Fermi-Dirac}

  \end{axis}
\end{tikzpicture}
\fi

%% file: figures/examples/datikz/3.tex
\ifdefined\getCaption%
Tree with aligned matrix. A probability tree with an aligned matrix listing the
possible outcomes, their probabilities and three columns for events described
in later tasks. It uses the grahdrawing library and requires LuaLaTeX.
\else\ifdefined\getSource%
\usetikzlibrary{graphs,graphdrawing,quotes,matrix,calc}
\usegdlibrary{trees}

\begin{tikzpicture}

  \graph [tree layout,
          grow=right,
          level distance=20mm,
          sibling distance=10mm,
          edge quotes={auto,
                       node font=\footnotesize,
                       inner sep=1pt,
                       outer sep=2pt,
                       shape=circle}
          ]
    {
    root [as=,inner sep=0pt] -> {
      NR [> "0.8"{',blue,draw}, as=$\bar{R}$] -> {
        T2 [> "0.9"{',blue,draw}, as=$T$] -> {
          NRTNZ [>"0.4"', as=$\bar{Z}$],
          NRTZ [>"0.6" {blue,draw}, as=$Z$]
        },
        D2 [> "0.1", as=$D$] -> {
          NRDNZ [>"0.5"', as=$\bar{Z}$],
          NRDZ [>"0.5", as=$Z$]
        }
      },
      R [> "0.2", as=$R$] -> {
        T1 [> "0.25"', as=$T$] -> {
          RTNZ [>"0.4"', as=$\bar{Z}$],
          RTZ [>"0.6" {blue,draw}, as=$Z$]
        },
        D1 [> "0.75" {blue,draw}, as=$D$] -> {
          RDNZ [>"0.9"', as=$\bar{Z}$],
          RDZ [>"0.1" {blue,draw}, as=$Z$]
        }
      }
    }
  };

  \matrix [matrix of math nodes,
           nodes in empty cells, 
           row sep={10mm,between origins}, 
           column sep=5mm,
           anchor=mat-2-1.center, 
           ]
  (mat) at ($(RDZ)+(2,0)$) 
  {
    \omega                & P(\omega)         & E_1     & E_2     & E_3     \\
    \{R;D;Z\}             & 0.015             & \bullet & \bullet &         \\
    \{R;D;\bar{Z}\}       & 0.135             & \bullet &         &         \\
    \{R;T;Z\}             & 0.03              & \bullet & \bullet &         \\
    \{R;T;\bar{Z}\}       & 0.02              & \bullet &         &         \\
    \{\bar{R};D;Z\}       & |[blue,draw]|0.04 &         & \bullet &         \\
    \{\bar{R};D;\bar{Z}\} & 0.04              &         &         & \bullet \\
    \{\bar{R};T;Z\}       & 0.432             & \bullet & \bullet &         \\
    \{\bar{R};T;\bar{Z}\} & 0.288             & \bullet &         &         \\
                          & 1                 &         &         &         \\
  };

  \foreach \x/\y in {1/2, 2/3, 3/4, 4/5, 5/6, 6/7, 7/8, 8/9}
    {\draw [-] ($(mat-\x-1.west -| mat-2-1.west)!0.5!(mat-\y-1.west)$) --
      ($(mat-\x-5.east -| mat-1-5.east)!0.5!(mat-\y-5.east -| mat-1-5.east)$);}

  \draw [double, shorten >=-1mm, shorten <=-1mm]
    ($(mat-9-2.west)!0.5!(mat-10-2.west)$) --
    ($(mat-9-2.east)!0.5!(mat-10-2.east)$);

  \node at (2.5,-4.5) {\small Given values are encircled.};

\end{tikzpicture}
\else
  \includegraphics{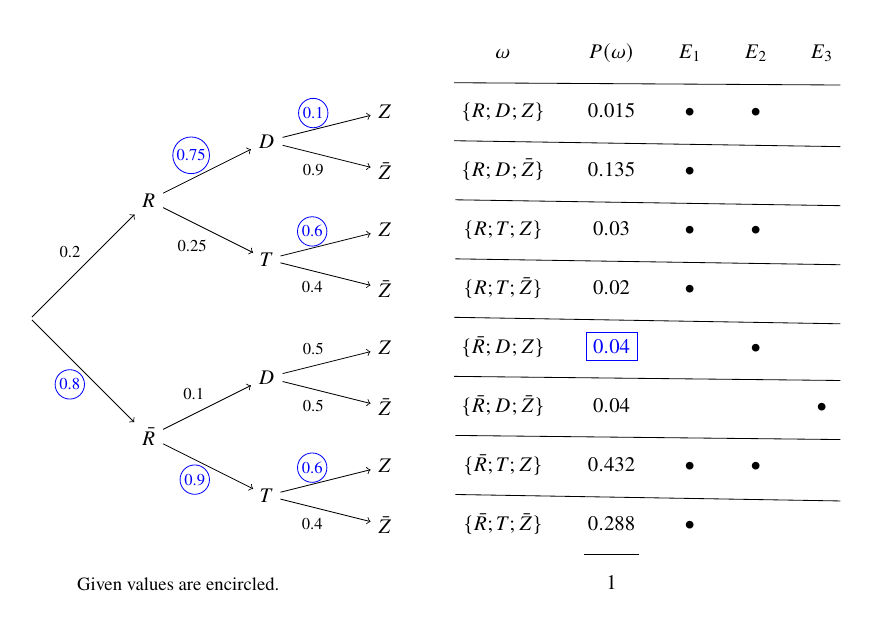}
\fi\fi

%% file: figures/examples/datikz/4.tex
\ifdefined\getCaption%
Our approach is a modified version of \textbf{meta-seq2seq}. A transformer
decoder (TD) is trained to produce a sequence of actions $a^Q_1, \ldots,
a^Q_{m}$ given a query instruction $I^Q$. The context are demonstrations $(I_k,
A_k)$ produced by our generative model. We use a transformer encoder-decoder
(T) to encode instructions and state $S$ and a transformer encoder (TE) to
encode actions. The transformers that process instructions (pink blocks)
receive state $S$ as the input of the encoder.
\else
\usetikzlibrary{calc}
\begin{tikzpicture}[auto, thick, node distance=6mm]
\draw
node at (0, 0) (A1) {$A_1$}
node [below of=A1] (I1) {$I_1$}
node [below of=I1] (dots1) {$...$}
node [below of=dots1] (An) {$A_n$}
node [below of=An] (In) {$I_n$}

node [draw, rectangle, fill=blue!10,rounded corners, right of=A1, minimum height=5mm, node distance=10mm] (At1) {TE}
node [draw, rectangle, fill=red!10,rounded corners, right of=I1, minimum height=5mm, node distance=10mm] (It1) {T}
node [draw, rectangle, fill=blue!10,rounded corners, right of=An, minimum height=5mm, node distance=10mm] (Atn) {TE}
node [draw, rectangle, fill=red!10,rounded corners, right of=In, minimum height=5mm, node distance=10mm] (Itn) {T}

node [draw, rectangle, fill=black!10,rounded corners, right of=dots1, minimum height=15mm, minimum width=30mm, anchor=south, node distance=33mm, rotate=90] (att) {attention}

node at ($(att.west) - (0, 6mm)$) [draw, rectangle, fill=red!10,rounded corners, minimum height=5mm] (Itq) {T}

node [below of=Itq, node distance=9mm] (Iq) {$I^Q$}

node at ($(att.south) + (10mm, 0mm)$) [draw, rectangle, fill=black!10,rounded corners, minimum height=30mm, minimum width=10mm, anchor=west] (TD) {TD}

node at ($(TD.north) + (0, 8mm)$) (outputs) {$a^Q_1, ..., a^Q_{m}$}

node at ($(TD.south) - (0, 8mm)$) (inputs) {SOS, $a^Q_1, ..., a^Q_{m-1}$}
;

\draw[->,>=stealth](A1) -- (At1);
\draw[->,>=stealth](I1) -- (It1);
\draw[->,>=stealth](An) -- (Atn);
\draw[->,>=stealth](In) -- (Itn);

\draw[->,>=stealth](At1) -- node[right,xshift=2mm] {$V_1$}(At1-|att.north);
\draw[->,>=stealth](It1) -- node[right,xshift=2mm] {$K_1$}(It1-|att.north);
\draw[->,>=stealth](Atn) -- node[right,xshift=2mm] {$V_n$}(Atn-|att.north);
\draw[->,>=stealth](Itn) -- node[right,xshift=2mm] {$K_n$}(Itn-|att.north);

\draw[->,>=stealth](Iq) -- (Itq);
\draw[->,>=stealth](Itq) -- (att);

\draw[->,>=stealth](inputs) -- (TD);
\draw[->,>=stealth](TD) -- (outputs);

\draw[->,>=stealth](att.south) -- (TD);

\end{tikzpicture}
\fi

%% file: figures/examples/copying/gpt-4/1.tex
\ifdefined\getCaption%
Motivating the definition of the  addition $\oplus$ and scalar multiplication
$\odot$ operations for $0 < \alpha <1$ in the Wasserstein optimal transport
space for transports $T_1, T_2$ (right), while in $\mathbb{R}^2$ optimal
transports are defined by vectors  $V_1, V_2$ (left).
\else
\begin{tikzpicture}
  \draw[->] (-1,0) -- (4,0) node[right] {$x$};
  \draw[->] (0,-1) -- (0,4) node[above] {$y$};
  \draw[->,blue] (0,0) -- (2,2) node[anchor=south east] {$V_1$};
  \draw[->,blue] (0,0) -- (3,1) node[anchor=west]  {$V_2$};
  \node at (1.5,-1) {In $\mathbb{R}^2$, optimal transports are defined by $V_1$ and $V_2$};
  \begin{scope}[xshift=6cm]
    \draw[->] (-1,0) -- (4,0) node[right] {$x'$};
    \draw[->] (0,-1) -- (0,4) node[above] {$y'$};
    \draw[blue] (2,2) circle (0.1cm) node[anchor=south west] {$T_1$};
    \draw[blue] (3,1) circle (0.1cm) node[anchor=south west] {$T_2$};
    \draw[->,red,very thick] (2,2) -- (3,1) node[midway,above]  {$+$};
    \draw[->,green,very thick] (2,2) -- (2,3) node[midway,right] {$\cdot \lambda$};
    \node at (1.5,-1) {In the Wasserstein space, $T_1$ and $T_2$ are added and scalar multiplied};
  \end{scope}
\end{tikzpicture}
\fi

%% file: figures/examples/copying/clima-13b/1.tex
\ifdefined\getCaption%
Motivating the definition of the  addition $\oplus$ and scalar multiplication
$\odot$ operations for $0 < \alpha <1$ in the Wasserstein optimal transport
space for transports $T_1, T_2$ (right), while in $\mathbb{R}^2$ optimal
transports are defined by vectors  $V_1, V_2$ (left).
\else
\begin{tikzpicture}[scale=1]
\fill[cyan] (0,0) -- (1,2) -- (1,1) -- cycle;
\draw[thick,->] (0,0) -- (1,0) node[below]{$v_1$};
\draw[thick,->] (0,0) -- (0,1) node[left]{$v_2$};
\draw[gray,thick,dashed] (0,0) -- (1,1);
\draw[thick,blue,->](0,0) -- (1,2) node[above]{$V_1$};
\draw[thick,blue,->] (0,0) -- (0.5,1) node[above]{$V_2$};

\fill[cyan] (3,0) -- (5,2) -- (6,1) -- cycle;
\draw[thick,->] (3,0) -- (5,0) node[below]{$T_1$};
\draw[thick,->] (3,0) -- (3,2) node[left]{$T_2$};
\draw[gray,thick,dashed] (3,0) -- (5,1);
\draw[thick,blue,->] (3,0) -- (5,2) node[above]{$T_1+T_2$};
\draw[thick,red,->] (3,0) -- (4,0.8) node[above]{$c_1 T_1$};
\draw[thick,red,->] (3,0) -- (3.8,0.5) node[above]{$c_2 T_2$};
\draw[thick,red,->] (3,0) -- (3.8,0.2) node[below]{$c_2 T_1$};
\draw[thick,blue,->] (3,0) -- (6,1) node[above]{$(T_1+T_2)/2$};
\end{tikzpicture}
\fi

%% file: figures/examples/copying/gpt-4/2.tex
\ifdefined\getCaption%
The influence of pre-training scale on the downstream performance. The value of
each method is the average metric values across all datasets in the privacy
protection setting with \textsc{LightGBM}.
\else
\begin{tikzpicture}
\begin{axis}[
  title={Influence of Pre-training Scale on Downstream Performance},
  xlabel={Pre-training Scale},
  ylabel={Downstream Performance},
  legend pos=north west,
  ymajorgrids=true,
  grid style=dashed,
  axis background/.style={fill=white}, 
]

\addplot[color=blue,mark=square,]
    coordinates {
    (1,1)(2,3)(3,2)(4,5)(5,7)
    };
    
\addplot[color=red,mark=triangle,]
    coordinates {
    (1,2)(2,2.5)(3,3)(4,6)(5,7)
    };
    
\addplot[color=green,mark=diamond,]
    coordinates {
    (1,1.5)(2,3.5)(3,4)(4,5)(5,6)
    };
    
\legend{Method 1, Method 2, Method 3}

\end{axis}
\end{tikzpicture}
\fi

%% file: figures/examples/copying/clima-13b/2.tex
\ifdefined\getCaption%
The influence of pre-training scale on the downstream performance. The value of
each method is the average metric values across all datasets in the privacy
protection setting with \textsc{LightGBM}.
\else
\begin{tikzpicture}
\begin{axis}[
    width=7cm,height=5cm,
    xlabel={pre-training scale},
    ylabel={downstream AP},
    xmin=0.5, xmax=5.5,
    ymin=52, ymax=55,
    xtick={1,2,3,4,5},
    xticklabels={1,10,100,1000,2000},
    ytick={52,53,54,55},
    legend pos=north east,
    ymajorgrids=true,
    grid style=dashed,
    axis background/.style={fill=white}, 
]

\addplot[
    color=blue,
    mark=square,
    ]
    coordinates {
    (1,53.22)(2,53.33)(3,53.50)(4,53.93)(5,54.14)
    };
    \legend{
    \textsc{InfoAcc}
    }
    
\addplot[
    color=red,
    mark=triangle,
    ]
    coordinates {
    (1,52.90)(2,53.04)(3,53.81)(4,54.07)(5,54.46)
    };
    \legend{
    \textsc{InfoGPT}
    }
    
\addplot[
    color=green,
    mark=star,
    ]
    coordinates {
    (1,53.85)(2,53.68)(3,54.16)(4,54.30)(5,54.58)
    };
    \legend{
    \textsc{InfoGPT-ViT}
    }
    
\addplot[
    color=orange,
    mark=otimes,
    ]
    coordinates {
    (1,54.20)(2,54.33)(3,54.81)(4,55.04)(5,55.33)
    };
    \legend{
    \textsc{InfoGPT-ViT$^+$}
    }
    
\end{axis}
\end{tikzpicture}
\fi

%% file: figures/examples/copying/gpt-4/3.tex
\ifdefined\getCaption%
An overlay \tikzname{} picture below the ordinary text, with the colored layer
appearing below the text rather than above it. Achieving this requires
adjusting the z-order of the \tikzname{} picture and the text.
\else\ifdefined\getSource%
\sffamily\Huge
\begin{tikzpicture}[overlay, remember picture]
    \node at (current page.center)
    {
        \begin{tikzpicture}
            \fill[red] (0,0) -- (5,7) -- (10,0) -- cycle;
        \end{tikzpicture}
    };
\end{tikzpicture}

\begin{center}
An overlay TikZ picture below the ordinary text, with the colored layer appearing below the text rather than above it. Achieving this requires adjusting the z-order of the TikZ picture and the text.
\end{center}
\else
  \includegraphics{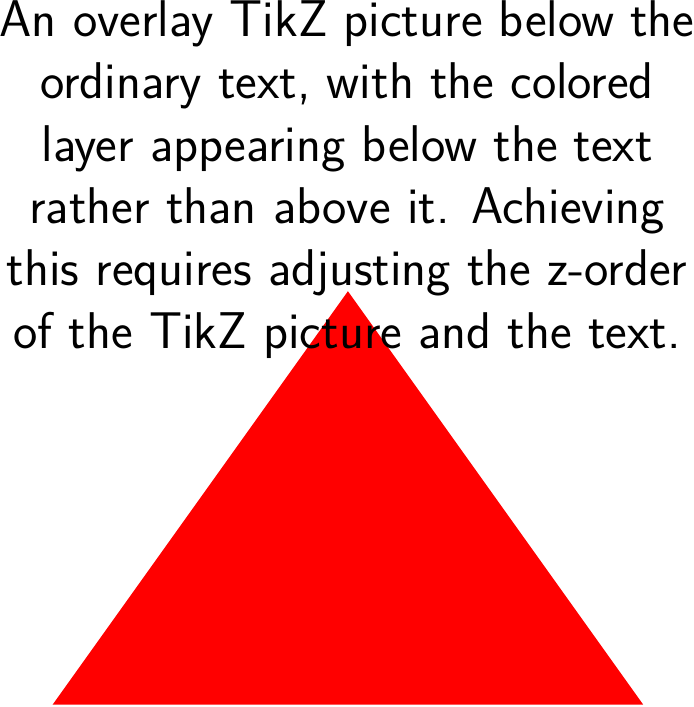}
\fi\fi

%% file: figures/examples/copying/clima-13b/3.tex
\ifdefined\getCaption%
An overlay \tikzname{} picture below the ordinary text, with the colored layer
appearing below the text rather than above it. Achieving this requires
adjusting the z-order of the \tikzname{} picture and the text.
\else\ifdefined\getSource%
\usetikzlibrary{tikzmark}
\newcommand{\sep}[1]{\hfill#1\hfill\relax}
\begin{minipage}[t]{.5\linewidth}
\begin{align*}
\sep{2+2=4}
\end{align*}
\end{minipage}%
\begin{minipage}[t]{.5\linewidth}
\tikzmarknode{2}{\rule{3cm}{3cm}}

\tikzmarknode{+}{\rule{3cm}{3cm}}

\tikzmarknode{3}{\rule{3cm}{3cm}}
\end{minipage}
\begin{tikzpicture}[overlay,remember picture]
\draw[red] (-1,-3) -- (5,-3) (-1,-2) -- (5,-2) (-1,-1) -- (5,-1)
(-1,0) -- (0,0) (1,0) -- (2,0) (3,0) -- (5,0)
(-1,1) -- (0,1) (1,1) -- (2,1) (3,1) -- (4,1) (5,1) -- (5.5,1)
(-1,2) -- (0,2) (1,2) -- (2,2) (3,2) -- (3.5,2) (4,2) -- (5,2)
(-1,3) -- (0,3) (1,3) -- (1.5,3) (2,3) -- (2.5,3) (3,3) -- (4,3)
(5,3) -- (5.5,3);
\end{tikzpicture}
\else
  \includegraphics{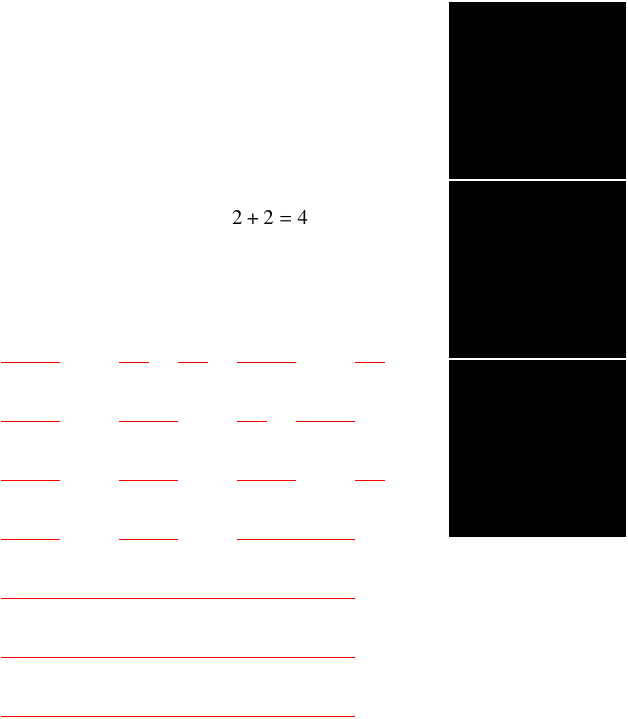}
\fi\fi